\documentclass{article}
\pdfoutput=1

 \PassOptionsToPackage{numbers, sort&compress, merge, square}{natbib}

\usepackage[preprint]{neurips_2018_mod}

\usepackage[utf8]{inputenc} 
\usepackage[T1]{fontenc}    
\usepackage[pdftex,
            pdfauthor={Alexandre N. Marques, Remi R. Lam, Karen E. Willcox},
            pdftitle={Contour location via entropy reduction leveraging multiple information sources},
            pdfkeywords={CLoVER, clover, contour entropy, Bayesian optimization, multi-fidelity optimization, multi-information source optimization, multi-fidelity adaptive learning, multi-information source adaptive learning, multi-fidelity classification, multi-information source classification, probability of excursion set, failure probability, multi-fidelity reliability analisys, multi-fidelity stability analysis},
            pdfproducer={Latex with hyperref},
            pdfcreator={pdflatex}]{hyperref}
\usepackage{url}            
\usepackage{booktabs}       
\usepackage{amsfonts}       
\usepackage{nicefrac}       
\usepackage{microtype}      
\usepackage{amsmath}
\usepackage{dsfont}
\usepackage{mathrsfs}
\usepackage{graphicx}
\usepackage{enumitem}

\title{Contour location via entropy reduction \\ leveraging multiple information sources}

\author{Alexandre N.~Marques \\
  Department of Aeronautics and Astronautics\\
  Massachusetts Institute of Technology\\
  Cambridge, MA 02139 \\
  \texttt{noll@mit.edu} \\
  \And
  Remi R.~Lam\\
  Center for Computational Engineering\\
  Massachusetts Institute of Technology\\
  Cambridge, MA 02139 \\
  \texttt{rlam@mit.edu}   
  \AND
  Karen E.~Willcox\\
  Institute for Computational Engineering and Sciences\\
  University of Texas at Austin\\
  Austin, TX 78712 \\
  \texttt{kwillcox@ices.utexas.edu}
}

\newcommand{\domain}{\ensuremath{\mathcal{D}}}
\newcommand{\zero}{\ensuremath{\mathcal{Z}}}
\newcommand{\x}{\ensuremath{\boldsymbol{x}}}
\newcommand{\zp}{\ensuremath{\mathbb{Z}^+}}
\DeclareMathOperator{\cov}{Cov}

\DeclareMathOperator{\lse}{\mathscr{H}}

\DeclareMathOperator{\maximize}{maximize}
\newcommand{\ce}{\ensuremath{c_{\epsilon}}}
\newcommand{\ord}[1]{\ensuremath{\mathcal{O}(#1)}}

\begin{document}

\maketitle


\begin{abstract}
 We introduce an algorithm to locate contours of functions
 that are expensive to evaluate.
 The problem of locating contours arises in many
 applications, including classification, constrained
 optimization, and  performance analysis of mechanical and
 dynamical systems (reliability, probability of failure,
 stability, etc.).
 Our algorithm locates contours using information from
 multiple sources, which are available in the form of
 relatively inexpensive, biased, and possibly noisy
 approximations to the original function.
 Considering multiple information sources can lead to
 significant cost savings.
 We also introduce the concept of \textit{contour entropy},
 a formal measure of uncertainty about the location of the
 zero contour of a function approximated by a statistical
 surrogate model.
 Our algorithm locates contours efficiently by maximizing
 the reduction of contour entropy per unit cost.
\end{abstract}


\section{Introduction}
\label{sec:intro}

In this paper we address the problem of locating contours of
functions that are expensive to evaluate.
This problem arises in several areas of science and
engineering.
For instance, in classification problems the contour
represents the boundary that divides objects of different
classes.
Another example is constrained optimization, where the
contour separates feasible and infeasible designs.
This problem also arises when analyzing the performance of
mechanical and dynamical systems, where contours divide
different behaviors such as stable/unstable, safe/fail, etc.
In many of these applications, function evaluations involve
costly computational simulations, or testing expensive
physical samples.
We consider the case when multiple information sources are
available, in the form of relatively inexpensive, biased,
and possibly noisy approximations to the original function.
Our goal is to use information from all available sources
to produce the best estimate of a contour under a fixed
budget.

We address this problem by introducing the CLoVER
(\textbf{C}ontour \textbf{Lo}cation \textbf{V}ia
\textbf{E}ntropy \textbf{R}eduction) algorithm.
CLoVER is based on a combination of principles from Bayesian
multi-information source optimization~\cite{lam:2015,
lam:2016, poloczek:2017} and information
theory~\cite{cover:2006}.
Our new contributions are:
\begin{itemize}[leftmargin=0.2in]
 \item The concept of \textit{contour entropy}, a measure of
       uncertainty about the location of the zero contour of
       a function approximated by a statistical surrogate
       model.
 \item An acquisition function that maximizes the reduction
       of contour entropy per unit cost.
 \item An algorithm that locates contours of functions
       using multiple information sources via reduction of
       contour entropy.
\end{itemize}

This work is related to the topic of Bayesian
multi-information source optimization
(MISO)~\cite{forrester:2007, swersky:2013, lam:2015,
lam:2016, poloczek:2017}.
Specifically, we use a statistical surrogate model to fit
the available data and estimate the correlation between
different information sources, and we choose the location
for new evaluations as the maximizer of an acquisition
function.
However, we solve a different problem than Bayesian 
optimization algorithms.
In the case of Bayesian optimization, the objective is to
locate the global maximum of an expensive-to-evaluate
function.
In contrast, we are interested in the entire set of points
that define a contour of the function.
This difference is reflected in our definition of an
acquisition function, which is fundamentally distinct from
Bayesian optimization algorithms.

Other algorithms address the problem of locating the contour
of expensive-to-evaluate functions, and are based on two
main techniques:
Support Vector Machine (SVM) and
Gaussian process (GP) surrogate.
CLoVER lies in the second category.

SVM~\cite{cortes:1995} is a commonly adopted classification
technique, and can be used to locate contours by defining
the regions separated by them as different classes.
Adaptive SVM~\cite{basudhar:2008, basudhar:2010,
lecerf:2015} and active learning with
SVM~\cite{schohn:2000, tong:2001, warmuth:2001} improve the
original SVM framework by adaptively selecting new samples
in ways that produce better classifiers with a smaller
number of observations.
Consequently, these variations are well suited for
situations involving expensive-to-evaluate functions.
Furthermore, Dribusch et al.~\cite{dribusch:2010} propose an
adaptive SVM construction that leverages multiple
information sources, as long as there is a predefined
fidelity hierarchy between the information sources.

Algorithms based on GP surrogates~\cite{bichon:2008,
ranjan:2008, picheny:2010, bect:2012, chevalier:2014,
wang:2016} use the uncertainty encoded in the surrogate to
make informed decisions about new evaluations, reducing
the overall number of function evaluations needed to locate
contours.
These algorithms differ mainly in the acquisition functions
that are optimized to select new evaluations.
Bichon et al.~\cite{bichon:2008},
Ranjan et al.~\cite{ranjan:2008}, and
Picheny et al.~\cite{picheny:2010} define acquisition
functions based on greedy reduction of heuristic measures
of uncertainty about the location of the contour,
whereas Bect et al.~\cite{bect:2012} and Chevalier et
al.~\cite{chevalier:2014} define acquisition functions
based on one-step look ahead reduction of quadratic loss
functions of the probability of an excursion set.
In addition, Stroh et al.~\cite{stroh:2017} use a GP
surrogate based on multiple information sources, under the
assumption that there is a predefined fidelity hierarchy
between the information sources.
Opposite to the algorithms discussed above,
Stroh et al.~\cite{stroh:2017} do not use the surrogate to
select samples.
Instead, a pre-determined nested LHS design allocates the
computational budget throughout the different information
sources.

CLoVER has two fundamental distinctions with respect to the
algorithms described above.
First, the acquisition function used in CLoVER is based on
one-step look ahead reduction of contour entropy, a formal
measure of uncertainty about the location of the contour.
Second, the multi-information source GP surrogate used in
CLoVER does not require any hierarchy between the
information sources.
We show that CLoVER outperforms the algorithms of
Refs.~\cite{bichon:2008, ranjan:2008, picheny:2010,
bect:2012, chevalier:2014, wang:2016} when applied to 
two problems involving a single information source.
One of these problems is discussed in
Sect.~\ref{sec:results}, while the other is discussed in the
supplementary material.

The remainder of this paper is organized as follows.
In Sect.~\ref{sec:problem} we present a formal problem
statement and introduce notation.
Then, in Sect.~\ref{sec:algorithm} we introduce the details
of the CLoVER algorithm, including the definition of the
concept of contour entropy.
Finally, in Sect.~\ref{sec:results} we present examples
that illustrate the performance of CLoVER.



\section{Problem statement and notation%
\protect\footnote{%
The statistical model used in the present algorithm is the
same introduced in~\cite{poloczek:2017}, and we attempt to
use a notation as similar as possible to this reference for
the sake of consistency.}
}
\label{sec:problem}

Let $g: \domain \mapsto \mathbb{R}$ denote a continuous
function on the compact set $\domain \in \mathbb{R}^d$, and
$g_{\ell}: \domain \mapsto \mathbb{R}$, $\ell \in [M]$,
denote a collection of the $M$ information sources (IS) that
provide possibly biased estimates of $g$.
(For $M \in \zp$, we use the notation
$[M] = \{1, \ldots, M\}$ and $[M]_0 = \{0, 1, \ldots, M\}$).
In general, we assume that observations of $g_{\ell}$ may be
noisy, such that they correspond to samples from the normal
distribution
$\mathcal{N}(g_{\ell}(\x), \lambda_{\ell}(\x))$.
We further assume that, for each IS $\ell$, the query cost
function, $c_{\ell}: \domain \mapsto \mathbb{R}^+$, and the
variance function $\lambda_{\ell}$ are known and
continuously differentiable over \domain\/.
Finally, we assume that $g$ can also be observed directly
without bias (but possibly with noise), and refer to it
as information source 0 (IS0), with query cost $c_0$ and
variance $\lambda_0$.

Our goal is to find the best approximation, within a fixed
budget, to a specific contour of $g$ by using a combination
of observations of $g_{\ell}$.
In the remainder of this paper we assume, without loss of
generality, that we are interested in locating the zero
contour of $g$, defined as the set
$\zero = \{z \in \domain \mid g(z) = 0\}$.




\section{The CLoVER algorithm}
\label{sec:algorithm}

In this section we present the details of the CLoVER
(\textbf{C}ontour \textbf{Lo}cation \textbf{V}ia
\textbf{E}ntropy \textbf{R}eduction) algorithm.
CLoVER has three main components:
(i)~a statistical surrogate model that combines information
    from all $M+1$ information sources, presented in
    Sect.~\ref{sub:model},
(ii)~a measure of the entropy associated with the zero
     contour of $g$ that can be computed from the surrogate,
     presented in Sect.~\ref{sub:entropy}, and
(iii)~an acquisition function that allows selecting
      evaluations that reduce this entropy measure,
      presented in Sect.~\ref{sub:acquisition}.
In Sect.~\ref{sub:hyper} we discuss the estimation of the
hyperparameters of the surrogate model, and in
Sect.~\ref{sub:summary} we show how these components are
combined to form an algorithm to locate the zero contour of
$g$.
We discuss the computational cost of CLoVER in the
supplementary material.

\subsection{Statistical surrogate model}
\label{sub:model}

CLoVER uses the statistical surrogate model introduced by
Poloczek et al.~\cite{poloczek:2017} in the context of
multi-information source optimization.
This model constructs a single Gaussian process (GP)
surrogate that approximates all information sources
$g_{\ell}$ simultaneously, encoding the correlations
between them.
Using a GP surrogate allows data assimilation using
standard tools of Gaussian process
regression~\cite{rasmussen:2005}.

We denote the surrogate model by $f$, with $f(\ell, \x)$
being the normal distribution that represents the belief
about IS $\ell$, $\ell \in [M]_0$, at location $\x$.
The construction of the surrogate follows from two modeling
choices:
(i)~a GP approximation to $g$ denoted by $f(0, \x)$, i.e.,
$f(0, \x) \sim GP(\mu_0, \Sigma_0)$, and
(ii)~independent GP approximations to the biases
$\delta_{\ell}(\x) = g_{\ell}(\x) - g(x)$,
$\delta_{\ell} \sim GP(\mu_{\ell}, \Sigma_{\ell})$.
Similarly to \cite{poloczek:2017}, we assume that
$\mu_{\ell}$ and $\Sigma_{\ell}$, $\ell \in [M]_0$, belong
to one of the standard parameterized classes of mean
functions and covariance kernels.
Finally, we construct the surrogate of $g_{\ell}$ as
$f(\ell, \x) = f(0, \x) + \delta_{\ell}(\x)$.
As a consequence, the surrogate model $f$ is a GP,
$f \sim GP(\mu, \Sigma)$, with
\begin{align}
 \mu(\ell, \x) &= \mathbb{E}[f(\ell, \x)] =
 \mu_0(\x) + \mu_{\ell}(\x), 
 \label{eq:mu} \\
 \Sigma \bigl( (\ell, \x), (m, \x') \bigr) &= 
 \cov \bigl( f(\ell, \x), f(m, \x') \bigr) = 
 \Sigma_0(\x, \x') +
 \mathds{1}_{\ell, m}\Sigma_{\ell}(\x, \x'),
 \label{eq:sigma}
\end{align}
where $\mathds{1}_{\ell, m}$ denotes the Kronecker's delta.

\subsection{Contour entropy}
\label{sub:entropy}

In information theory~\cite{cover:2006}, the concept of
entropy is a measure of the uncertainty in the outcome of a
random process.
In the case of a discrete random variable $W$ with $k$
distinct possible values $w_i$, $i \in [k]$, entropy is
defined by
\begin{equation}
 H(W) = -\sum_{i=1}^k P(w_i) \ln P(w_i),
\end{equation}
where $P(w_i)$ denotes the probability mass of value $w_i$.
It follows from this definition that lower values of entropy
are associated to processes with little uncertainty
($P(w_i) \approx 1$ for one of the possible outcomes).

\begin{figure}[t!]
 \begin{center}
  \includegraphics[scale=0.4, 
                   trim={0, 0.1in, 0, 0.3in}, 
                   clip]{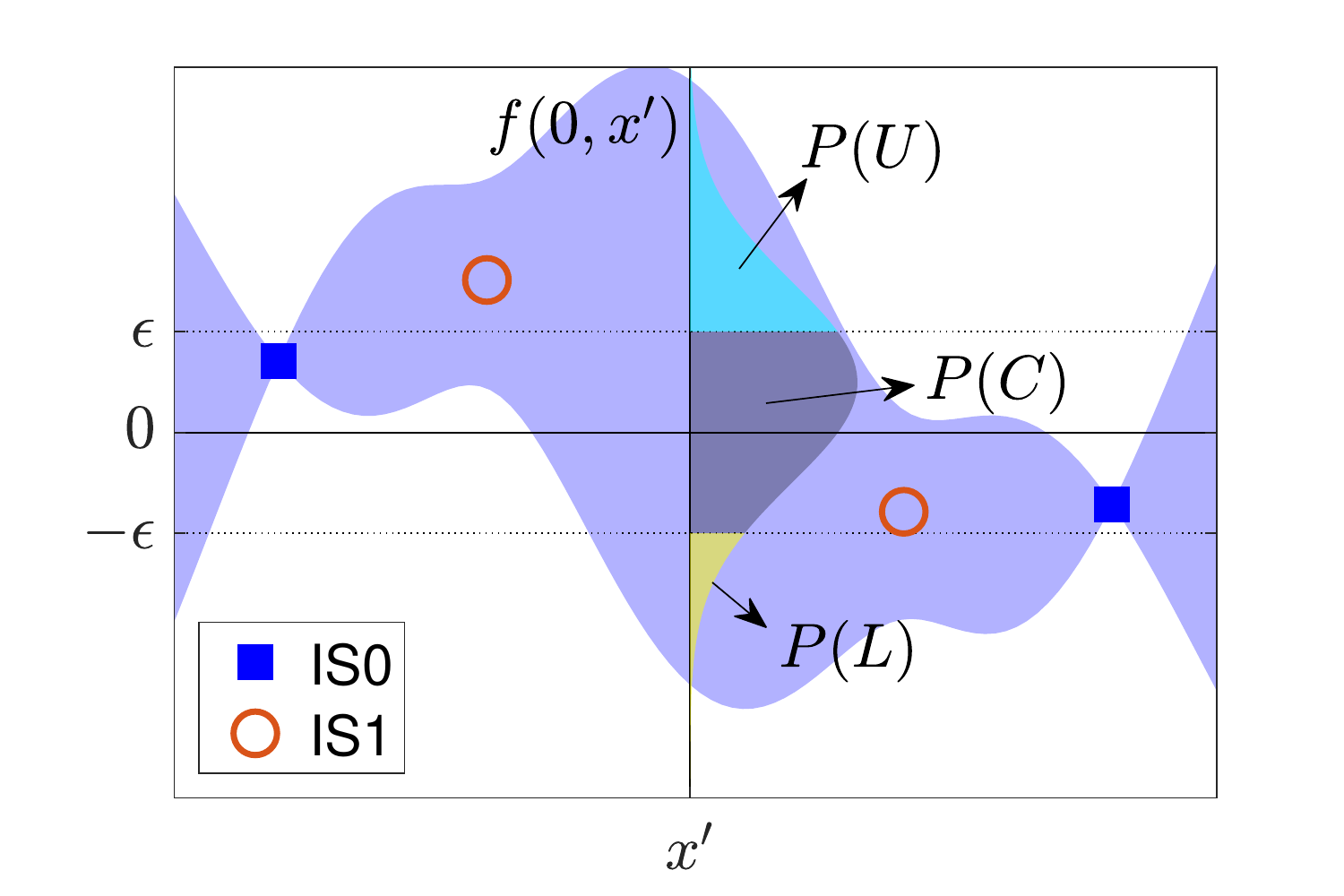}
  \includegraphics[scale=0.33, 
                   trim={0, 0.25in, 0, 0.05in}, 
                   clip]{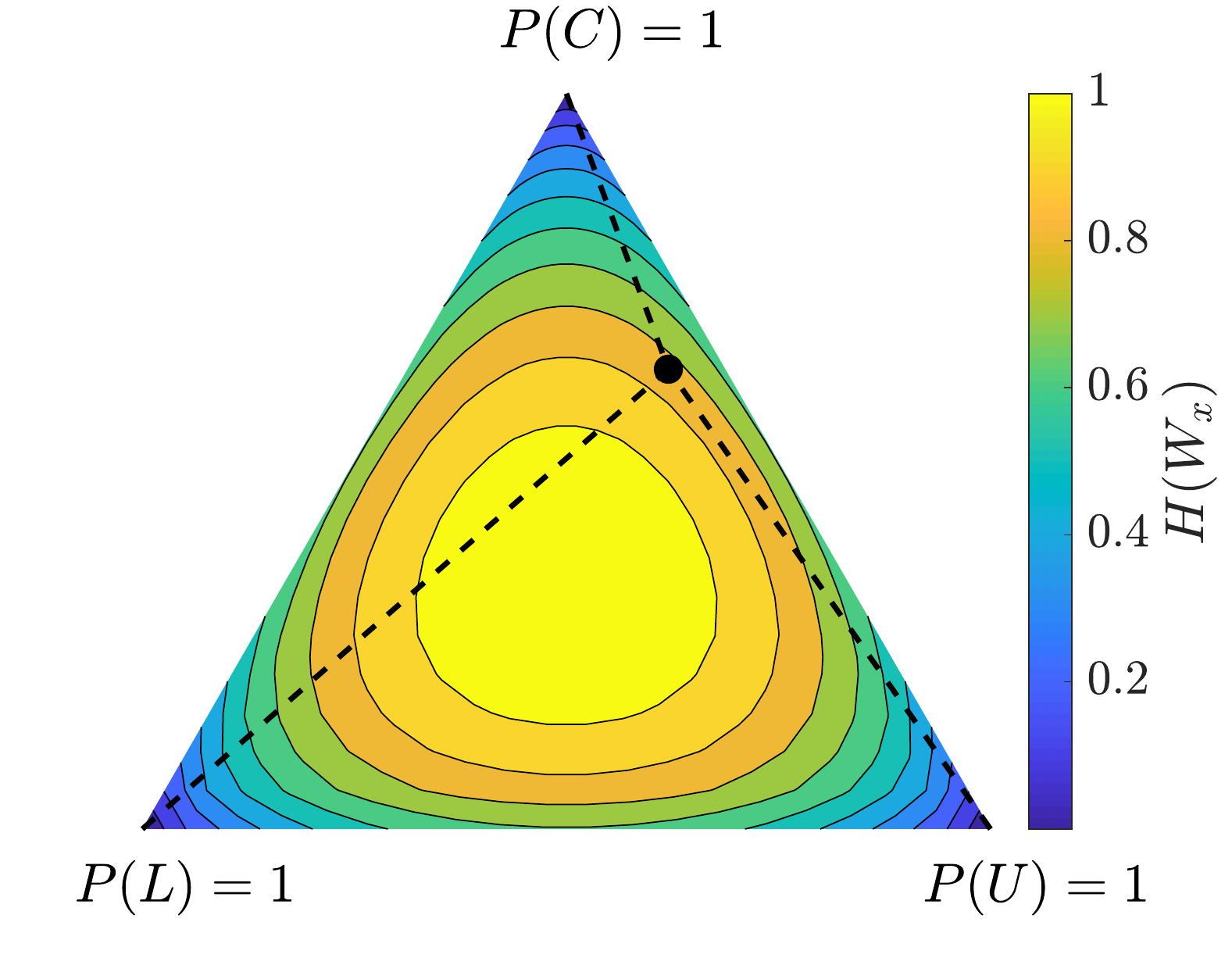}
 \end{center}
 \caption{Left: GP surrogate, distribution $f(0, x')$ and
                probability mass of events $L$, $C$, and
                $U$, which define the random variable
                $W_{x'}$.
          Right: Entropy $H(W_x)$ as a function of the
                 probability masses.
                 The black dot corresponds to $H(W_{x'})$.}
 \label{fig:Wx}
\end{figure}

We introduce the concept of \textit{contour entropy} as the
entropy of a discrete random variable associated with the
uncertainty about the location of the zero contour of $g$,
as follows.
For any given $\x \in \domain$, the posterior distribution
of $f(0, \x)$ (surrogate model of $g(\x)$), conditioned on
all the available evaluations, is a normal random variable
with known mean $\mu(0, \x)$ and variance $\sigma^2(0, \x)$.
Given $\epsilon(\x) \in \mathbb{R}^+$, an observation $y$ of
this random variable can be classified as one of the
following three events:
$y < -\epsilon(\x)$ (denoted as event $L$),
$|y| < \epsilon(\x)$ (denoted as event $C$), or
$y > \epsilon(\x)$ (denoted as event $U$).
These three events define a discrete random variable,
$W_{\x}$, with probability mass
$P(L) = \Phi((-\mu(0, \x) -\epsilon(\x))/\sigma(0, \x))$,
$P(C) = \Phi((-\mu(0, \x) +\epsilon(\x))/\sigma(0, \x)) 
      - \Phi((-\mu(0, \x) -\epsilon(\x))/\sigma(0, \x))$,
$P(U) = \Phi((\mu(0, \x) - \epsilon(\x))/\sigma(0, \x))$,
where $\Phi$ is the unit normal cumulative distribution
function.
Figure~\ref{fig:Wx} illustrates events $L$, $C$, and $U$,
and the probability mass associated with each of them.
In particular, $P(C)$ measures the probability of
$g(\x)$ being within a band of width $2 \epsilon(\x)$
surrounding the zero contour, as estimated by the GP
surrogate.
The parameter $\epsilon(\x)$ represents a tolerance in our
definition of a zero contour.
As the algorithm gains confidence in its predictions, it is
natural to reduce $\epsilon(\x)$ to tighten the bounds on the
location of the zero contour.
As discussed in the supplementary material, numerical
experiments indicate that $\epsilon(\x) = 2 \sigma(\x)$
results in a good balance between exploration and
exploitation.

The entropy of $W_{\x}$ measures the uncertainty in whether
$g(\x)$ lies below, within, or above the tolerance
$\epsilon(\x)$, and is given by
\begin{equation}
 \label{eq:HWx}
  H(W_{\x}; f) = -P(L)\log P(L) - P(C)\log P(C) -P(U)\log P(U).
\end{equation}
This entropy measures uncertainty at parameter value $\x$ only.
To characterize the uncertainty of the location of the zero
contour, we define the \textit{contour entropy} as
\begin{equation}
 \label{eq:lse}
 \lse(f) = \dfrac{1}{V(\domain)}
  \int_{\domain} H(W_{\x}; f)\, d\x,
\end{equation}
where $V(\domain)$ denotes the volume of $\domain$.

\subsection{Acquisition function}
\label{sub:acquisition}

CLoVER locates the zero contour by selecting samples that
are likely reduce the contour entropy at each new iteration.
In general, samples from IS$0$ are the most informative
about the zero contour of $g$, and thus are more likely to
reduce the contour entropy, but they are also the most
expensive to evaluate.
Hence, to take advantage of the other $M$ IS available, the
algorithm performs observations that maximize the expected
reduction in contour entropy, normalized by the query cost.

Consider the algorithm after $n$ samples evaluated at
$X_n = \{ ( \ell^i, \x^i ) \}_{i=1}^n$, which result in
observations $Y_n = \{ y^i \}_{i=1}^n$.
We denote the posterior GP of $f$, conditioned on
$\{ X_n, Y_n \}$, as $f^n$, with mean $\mu^n$ and covariance
matrix $\Sigma^n$.
Then, the algorithm selects a new parameter value
$\x \in \domain$, and IS $\ell \in [M]_0$ that satisfy the
following optimization problem.
\begin{equation}
 \label{eq:optimization}
  \underset{{\ell \in [M]_0,\, \x \in \domain}}{\maximize}
   \quad u(\ell, \x; f^n),
\end{equation}
where
\begin{equation}
 \label{eq:acquisition}
 u(\ell, \x; f^n) =
 \dfrac
  { \mathbb{E}_y
   [ \lse(f^n) -
     \lse(f^{n+1}) \mid 
      \ell^{n+1} = \ell, \, \x^{n+1} = \x ] }
  {c_{\ell}(\x)},
\end{equation}
and the expectation is taken over the distribution of
possible observations, \linebreak
$y^{n+1} \sim \mathcal{N} \bigl( \mu^n(\ell, \x),
\Sigma^n  ((\ell, \x), (\ell, \x)) \bigr)$.
To make the optimization problem tractable, the search
domain is replaced by a discrete set of points
$\mathcal{A} \subset \domain$, e.g., a Latin Hypercube
design. 
We discuss how to evaluate the acquisition function $u$
next.

Given that $f^n$ is known, $\lse(f^n)$ is a deterministic
quantity that can be evaluated
from~(\ref{eq:HWx}--\ref{eq:lse}).
Namely, $H(W_x; f^n)$ follows directly from \eqref{eq:HWx},
and the integration over \domain\/ is computed via a Monte
Carlo-based approach (or regular quadrature if the
dimension of \domain\/ is relatively small).

Evaluating $\mathbb{E}_y [\lse(f^{n+1})]$ requires a few
additional steps.
First, the expectation operator commutes with the
integration over \domain\/.
Second, for any $\x' \in \domain$, the entropy
$H(W_{\x'}; f^{n+1})$ depends on $y^{n+1}$ through its
effect on the mean $\mu^{n+1}(0, \x')$ (the covariance
matrix $\Sigma^{n+1}$ depends only on the location of the
samples).
The mean is affine with respect to the observation $y^{n+1}$
and thus is distributed normally:
$\mu^{n+1}(0, \x') \sim \mathcal{N}(\mu^n(0, \x'),
\bar{\sigma}^2(\x'; \ell, \x))$,
where $\bar{\sigma}^2(\x'; \ell, \x) =
\bigl( \Sigma^n ( (0,\x'), (\ell, \x) ) \bigr)^2 /
\bigl( \lambda_{\ell}(\x)
+ \Sigma^n ( (\ell,\x), (\ell, \x) ) \bigr)$.
Hence, after commuting with the integration over \domain\/,
the expectation with respect to the distribution of
$y^{n+1}$ can be equivalently replaced by the expectation
with respect to the distribution of $\mu^{n+1}(0, \x')$,
denoted by $\mathbb{E}_{\mu}[(.)]$.

\begin{figure}[t!]
 \begin{center}
  \includegraphics[scale=0.3,
                   trim={0, 0.1in, 0, 0.3in},
                   clip]{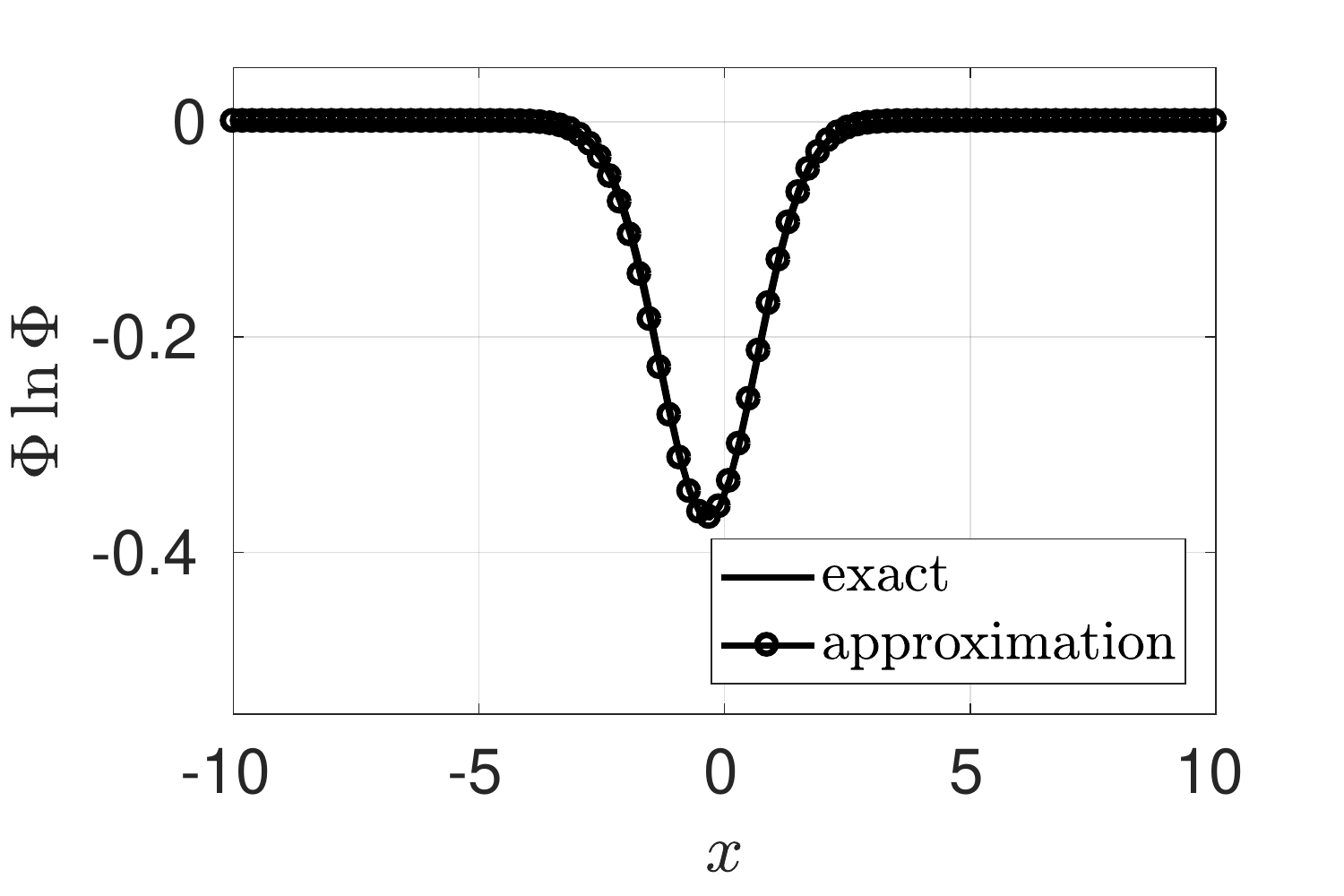}
  \hspace{0.2in}
  \includegraphics[scale=0.3,
                   trim={0, 0.1in, 0, 0.3in},
                   clip]{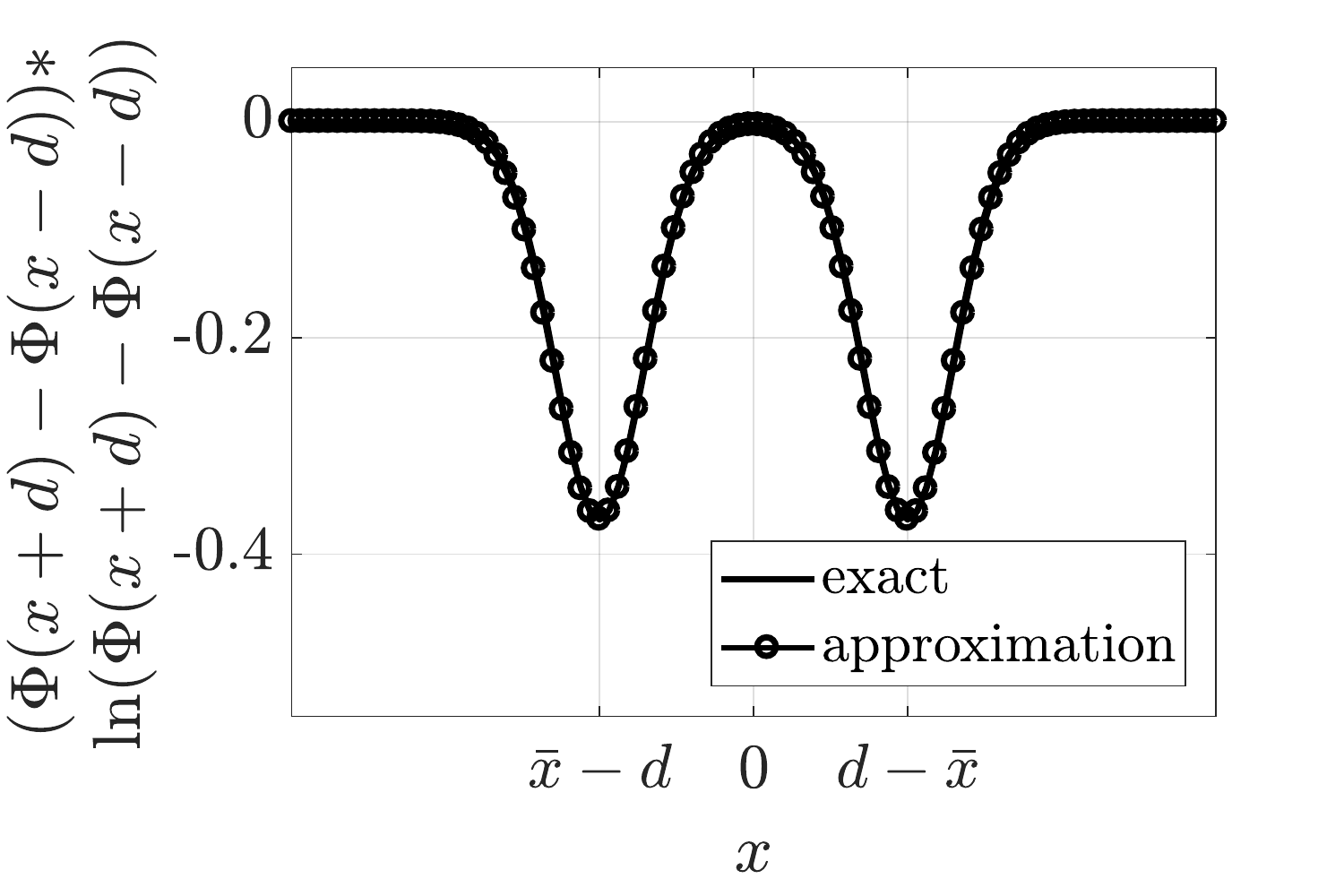}
 \end{center}
 \caption{Comparison between functions involving products of
          $\Phi$ and $\ln \Phi$ and approximations
          (\ref{eq:approximation_m}--\ref{eq:approximation_d}).}
 \label{fig:approximations}
\end{figure}

Third, in order to compute the expectation operator
analytically, we introduce the following approximations.
\begin{align}
 &\Phi(x) \ln \Phi(x) \approx \sqrt{2 \pi} \, c \varphi(x - \bar{x}),
 \label{eq:approximation_m} \\
 &(\Phi(x+d) - \Phi(x-d))
 \ln (\Phi(x+d) - \Phi(x-d)) \approx
 \sqrt{2 \pi} \, c \bigl( \varphi(x - d + \bar{x}) 
  + \varphi(x + d - \bar{x}) \bigr),
 \label{eq:approximation_d}
\end{align}
where $\varphi$ is the normal probability density function,
$\bar{x} = \Phi^{-1}(e^{-1})$, and 
$c = \Phi(\bar{x}) \ln \Phi(\bar{x})$.
Figure~\ref{fig:approximations} shows the quality of these
approximations.
Then, we can finally write
\begin{equation}
 \begin{split}
  &\mathbb{E}_y
  [ \lse(f^{n+1}) \mid \ell^{n+1} = \ell,\, \x^{n+1} = \x ] \\
  &= \dfrac{1}{V(\domain)} \int_{\domain}
     \mathbb{E}_{\mu} 
     [ H(W_{\x'}; f^{n+1}) |
       \ell^{n+1} = \ell, \x^{n+1} = \x ] \, d\x' \\
  &\approx -\dfrac{c}{V(\domain)} \int_{\domain}
     r_{\sigma}(\x; \ell, \x)
     \sum_{i=0}^1 \sum_{j=0}^1
     \exp \left( -\dfrac{1}{2} \left(
      \dfrac{\mu^n(0, \x') + (-1)^i \epsilon}
            {\hat{\sigma}(\x'; \ell, \x)} 
      + (-1)^j \bar{x} r_{\sigma}(\x'; \ell, \x)
      \right)^2\right) \, d\x',
 \end{split}
\end{equation}
where
\begin{align*}
 \hat{\sigma}^2(\x'; \ell, \x) &=
 \Sigma^{n+1} ( (0, \x'),(0, \x') )
 + \bar{\sigma}^2(\x'; \ell, \x),
 &
 r_{\sigma}^2(\x'; \ell, \x) = 
 \dfrac
 { \Sigma^{n+1} ( (0, \x'),(0, \x') ) }
 { \hat{\sigma}^2(\x'; \ell, \x) }.
\end{align*}

\subsection{Estimating hyperparameters}
\label{sub:hyper}

Our experience indicates that the most suitable approach to
estimate the hyperparameters depends on the problem.
Maximum a posteriori (MAP) estimates normally perform
well if reasonable guesses are available for the priors of
hyperparameters.
On the other hand, maximum likelihood estimates (MLE)
may be sensitive to the randomness of the initial data,
and normally require a larger number of evaluations to
yield appropriate results.

Given the challenge of estimating hyperparameters with small
amounts of data, we recommend updating these estimates
throughout the evolution of the algorithm.
We adopt the strategy of estimating the hyperparameters
whenever the algorithm makes a new evaluation of IS0.
The data obtained by evaluating IS0 is used directly to
estimate the hyperparameters of $\mu_0$ and $\Sigma_0$.
To estimate the hyperparameters of $\mu_{\ell}$ and
$\Sigma_{\ell}$, $\ell \in [M]$, we evaluate all other $M$
information sources at the same location and compute the
biases $\delta_{\ell} = y_{\ell} - y_0$, where $y_{\ell}$
denotes data obtained by evaluating IS $\ell$.
The biases are then used to estimate the hyperparameters of
$\mu_{\ell}$ and $\Sigma_{\ell}$.

\subsection{Summary of algorithm}
\label{sub:summary}

\begin{enumerate}[leftmargin=0.2in]
 \item Compute an initial set of samples by evaluating all
       $M+1$ IS at the same values of $\x \in \domain$.
       Use samples to compute hyperparameters and the
       posterior of $f$.
 \item Prescribe a set of points
       $\mathcal{A} \subset \domain$ which will be used as
       possible candidates for sampling.
 \item Until budget is exhausted, do:
  \begin{enumerate}
   \item Determine the next sample by solving the
         optimization problem~\eqref{eq:optimization}.
   \item Evaluate the next sample at location $x^{n+1}$
         using IS $\ell^{n+1}$.
   \item Update hyperparameters and posterior of $f$.
  \end{enumerate}
 \item Return the zero contour of $\mathbb{E}[f(0, \x)]$.
\end{enumerate}


\section{Numerical results}
\label{sec:results}

In this section we present three examples that demonstrate
the performance of CLoVER.
The first two examples involve multiple information sources,
and illustrate the reduction in computational cost that can
be achieved by combining information from multiple sources
in a principled way.
The last example compares the performance of CLoVER to that
of competing GP-based algorithms, showing that CLoVER can
outperform existing alternatives even in the case of a
single information source.

\subsection{Multimodal function}
\label{sub:multimodal}

In this example we locate the zero contour of the following
function within the domain \linebreak
$\domain = [-4, 7] \times [-3, 8]$.
\begin{equation}
 \label{eq:ex1}
 g(\x) = \dfrac{(x_1^2 + 4)(x_2 - 1)}{20}
       - \sin \left( \dfrac{5 x_1}{2} \right) - 2.
\end{equation}
This example was introduced in Ref.~\cite{bichon:2008} in
the context of reliability analysis, where the zero contour
represents a failure boundary.
We explore this example further in the supplementary
material, where we compare CLoVER to competing algorithms
in the case of a single information source.
To demonstrate the performance of CLoVER in the presence of
multiple information sources, we introduce the following
biased estimates of $g$:
\begin{align*}
 g_1(\x) &= g(\x)
   + \sin \left( \dfrac{5}{22}
       \left (x_1 + \dfrac{x_2}{2} \right) 
         + \dfrac{5}{4}\right), &
 g_2(\x) &= g(\x)
   + 3 \sin \left( \dfrac{5}{11} (x_1 + x_2 + 7) \right).
\end{align*}
We assume that the query cost of each information source is
constant: $c_0 = 1$, $c_1 = 0.01$, $c_2 = 0.001$.
We further assume that all information sources can be
observed without noise.

\begin{figure}[t!]
 \begin{center}
  \includegraphics[scale=0.4,
                   trim={0 1.85in 0 0.05in},
                   clip]{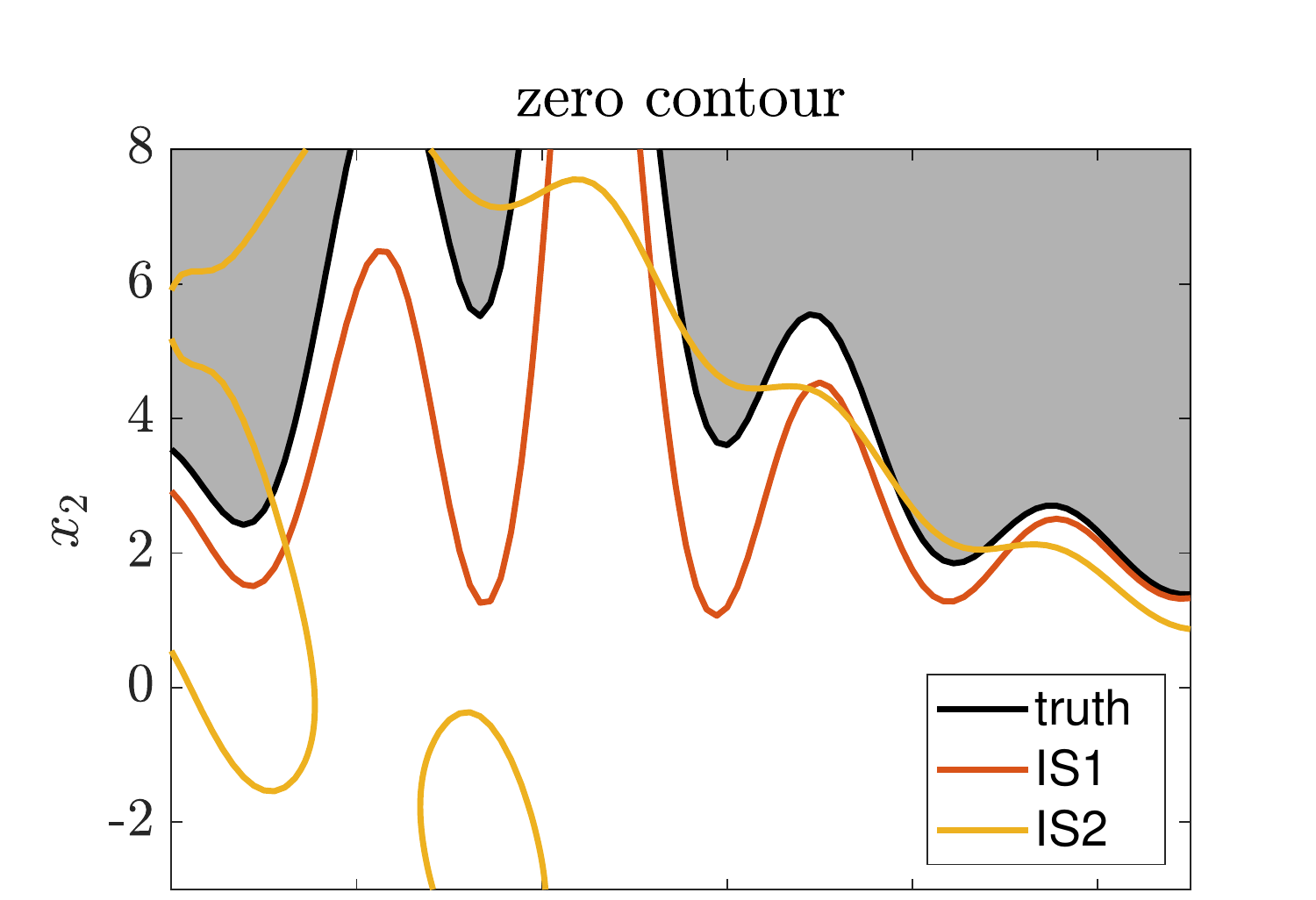}
  \includegraphics[scale=0.4,
                   trim={0.55in 1.85in 0 0.05in},
                   clip]{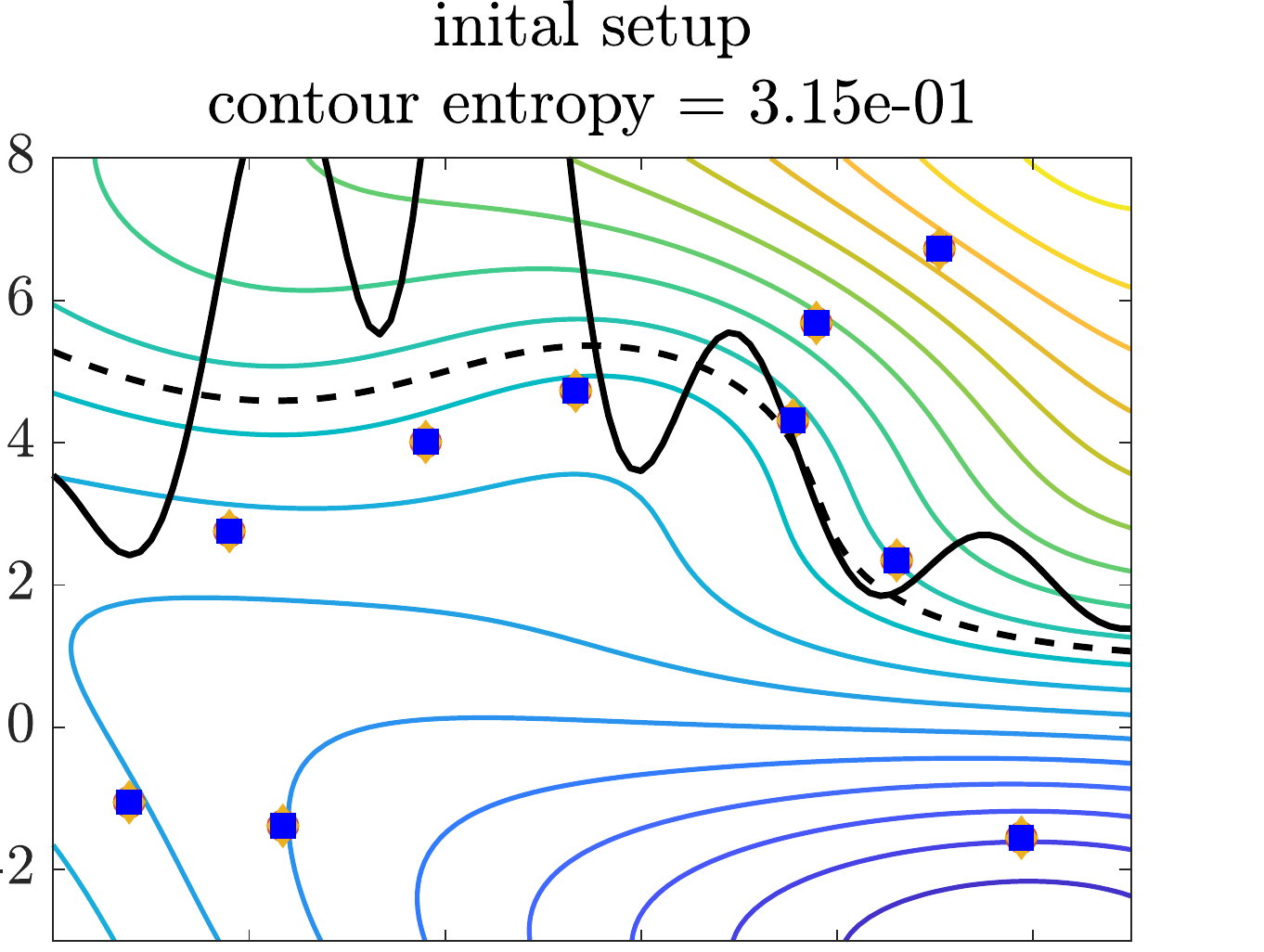} \\[5pt]
  \includegraphics[scale=0.4,
                   trim={0 0.5in 0 0.05in},
                   clip]{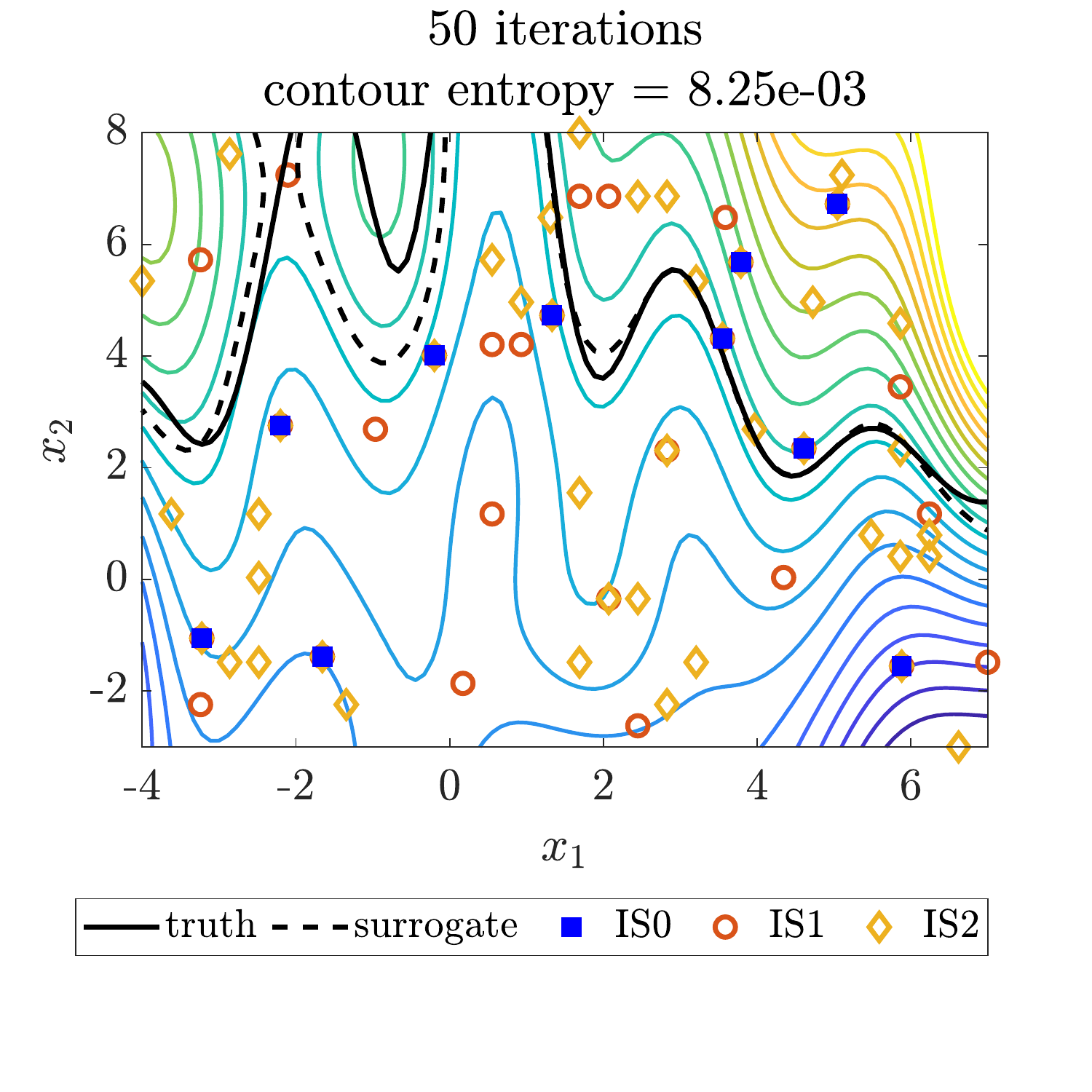}
  \includegraphics[scale=0.4,
                   trim={0.55in 0.5in 0 0.05in},
                   clip]{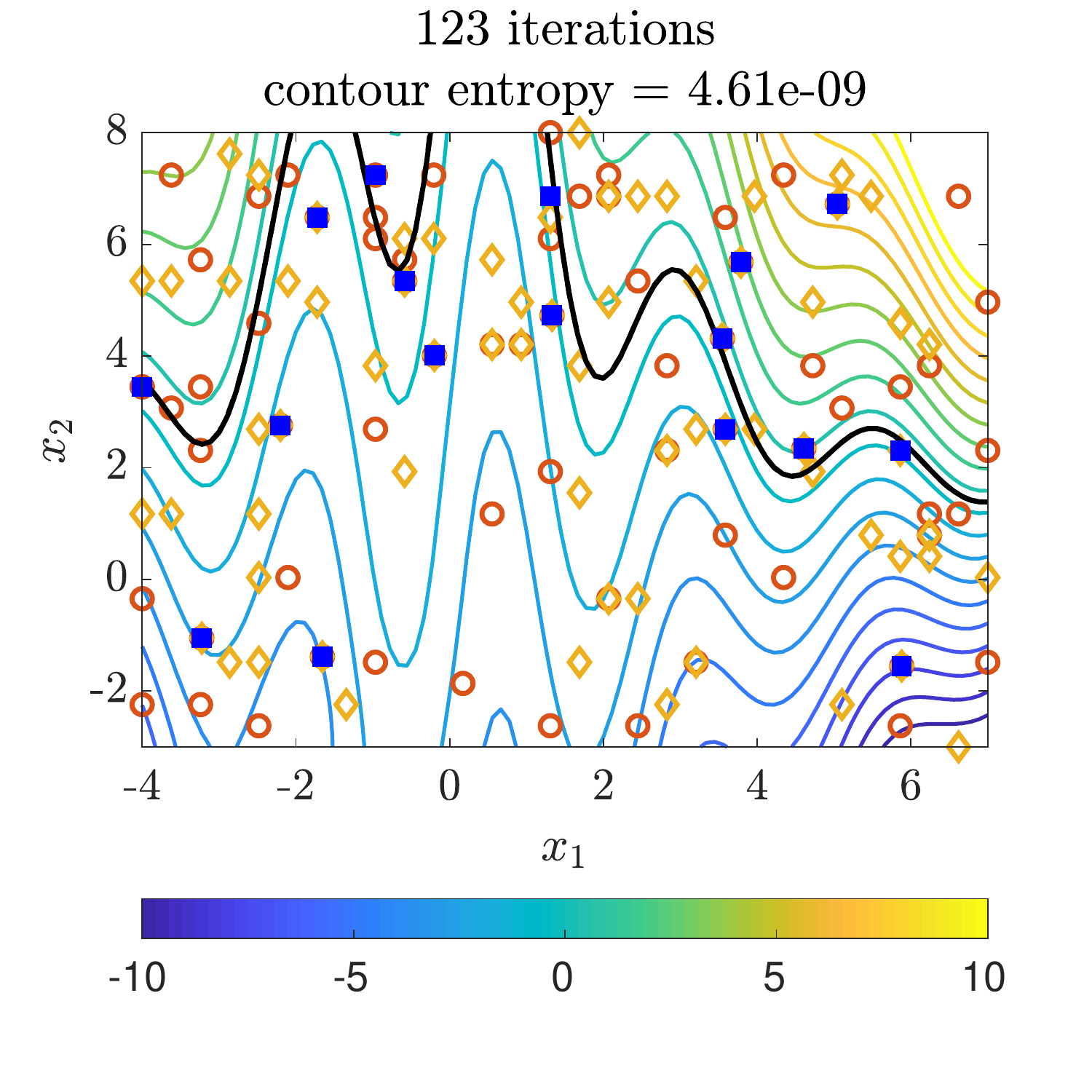}
 \end{center}
 \caption{Locating the zero contour of the multimodal
          function \eqref{eq:ex1}.
          Upper left: Zero contour of IS0, IS1, and IS2.
          Other frames: Samples and predictions made by
          CLoVER at several iterations.
          Dashed black line: Zero contour predicted by the
          surrogate model.
          Colors: Mean of the surrogate model $f(0, \x)$.
          CLoVER obtains a good approximation to the zero
          contour with only 17 evaluations of expensive
          IS0.}
 \label{fig:ex1:snapshots}
\end{figure}

Figure~\ref{fig:ex1:snapshots} shows predictions made by
CLoVER at several iterations of the algorithm.
CLoVER starts with evaluations of all three IS at the same
10 random locations.
These evaluations are used to compute the hyperparameters
using MLE, and to construct the surrogate model.
The surrogate model is based on zero mean functions and
squared exponential covariance
kernels~\cite{rasmussen:2005}.
The contour entropy of the initial setup is $\lse = 0.315$,
which indicates that there is considerable uncertainty in
the estimate of the zero contour.
CLoVER proceeds by exploring the parameter space using
mostly IS2, which is the model with the lowest query cost.
The algorithm stops after 123 iterations, achieving a
contour entropy of $\lse = 4 \times 10^{-9}$.
Considering the samples used in the initial setup, CLoVER
makes a total of 17 evaluations of IS0, 68 evaluations of
IS1, and 68 evaluations of IS2.
The total query cost is 17.8.
We repeat the calculations 100 times using different
values for the initial 10 random evaluations, and the 
median query cost is 18.1.
In contrast, the median query cost using a single
information source (IS0) is 38.0, as shown in the
supplementary material.
Furthermore, at query cost 18.0, the median contour
entropy using a single information source is
$\lse = 0.19$.

\begin{figure}[t!]
 \begin{center}
  \includegraphics[scale=0.4,
                   trim={0, 0, 0, 0.15in},
                   clip]{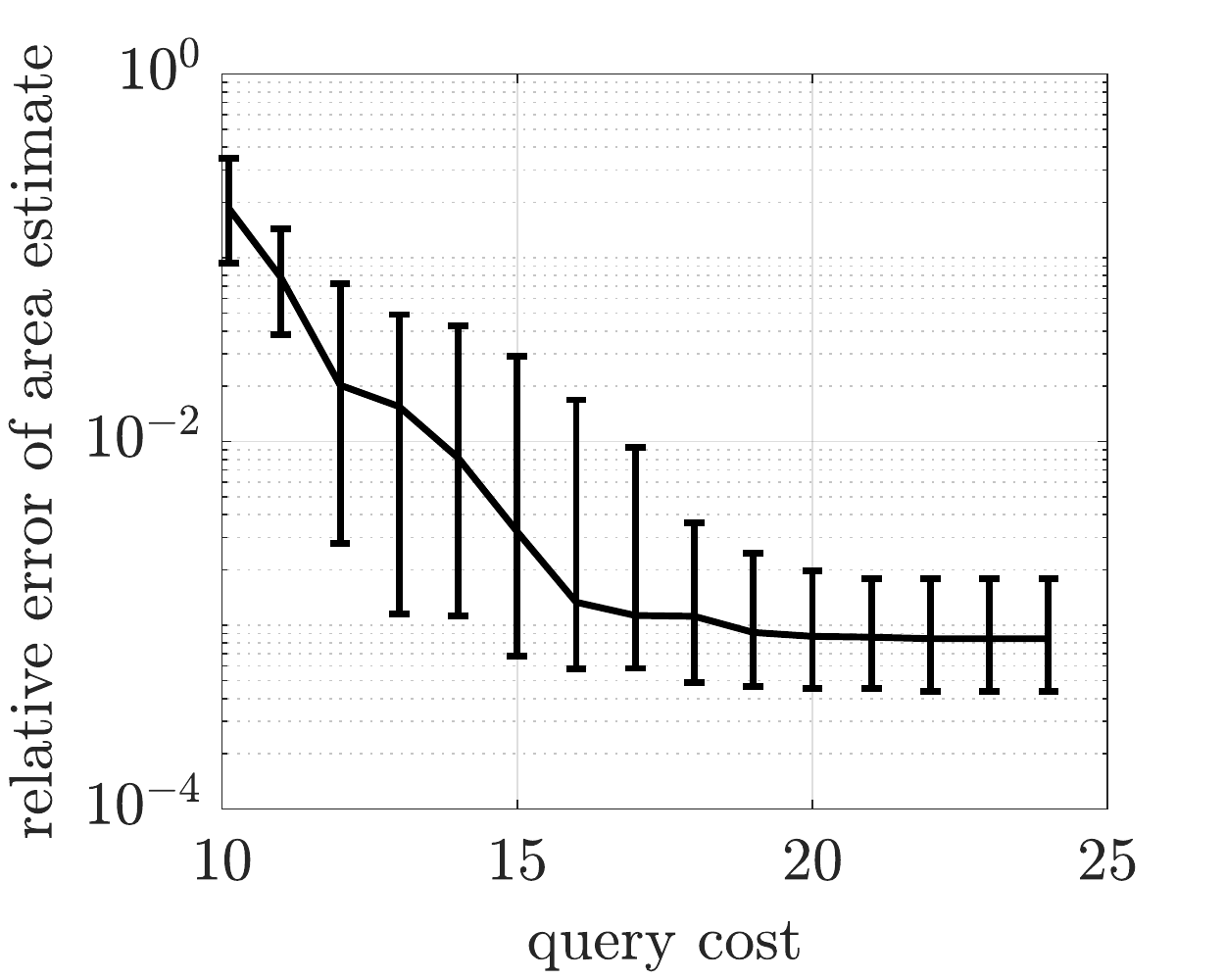}
  \includegraphics[scale=0.4,
                   trim={0, 0, 0, 0.15in},
                   clip]{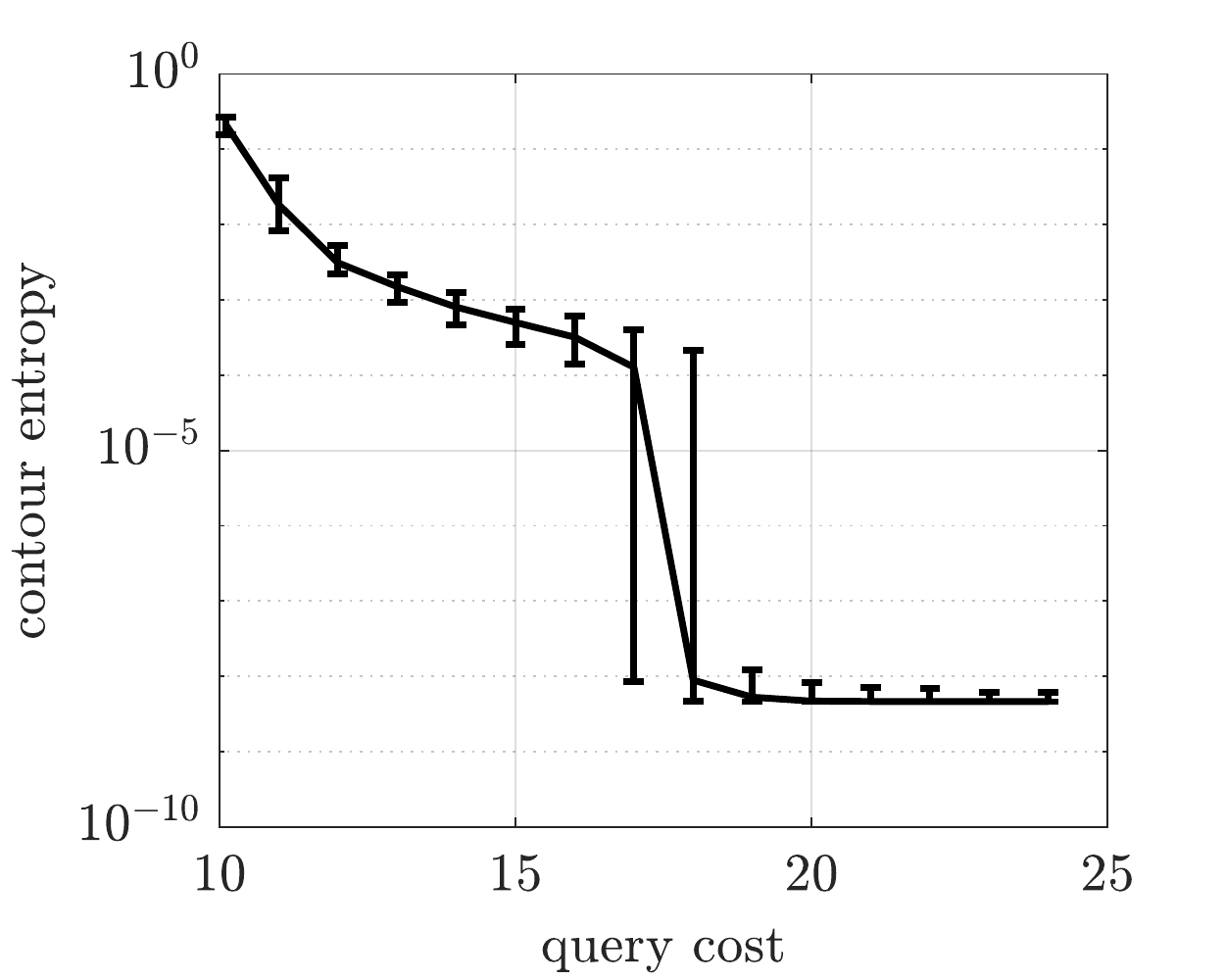}
 \end{center}
 \caption{Left: Relative error in the estimate of the area
                of the set $S$.
          Right: Contour entropy.
          Median, 25, and 75 percentiles.}
 \label{fig:ex1:convergence}
\end{figure}

We assess the accuracy of the zero contour estimate produced
by CLoVER by measuring the area of the set
$S = \{ \x \in \domain \mid g(\x) > 0\}$ (shaded region
shown on the top left frame of
Figure~\ref{fig:ex1:snapshots}).
We estimate the area using Monte Carlo integration with
$10^6$ samples in the region $[-4, 7] \times [1.4, 8]$.
We compute a reference value by averaging 20 Monte Carlo
estimates based on evaluations of $g$:
$\text{area}(S) = 36.5541$.
Figure~\ref{fig:ex1:convergence} shows the relative error in
the area estimate obtained with 100 evaluations of CLoVER.
This figure also shows the evolution of the contour entropy.

\subsection{Stability of tubular reactor}
\label{sub:tubular}

We use CLoVER to locate the stability boundary of a
nonadiabatic tubular reactor with a mixture of two chemical
species.
This problem is representative of the operation of
industrial chemical reactors, and has been the subject of
several investigations, e.g.~\cite{heinemann:1981}.
The reaction between the species releases heat, increasing
the temperature of the mixture.
In turn, higher temperature leads to a nonlinear increase in
the reaction rate.
These effects, combined with heat diffusion and convection,
result in complex dynamical behavior that can lead to
self-excited instabilities.
We use the dynamical model described in
Refs.~\cite{zhou:2010, peherstorfer:2016}.
This model undergoes a H\"opf bifurcation, when the response
of the system transitions from decaying oscillations to
limit cycle oscillations.
This transition is controlled by the Damk\"ohler number $D$,
and here we consider variations in the range
$D \in [0.16, 0.17]$ (the bifurcation occurs at the critical
Damk\"ohler number $D_{cr} = 0.165$).
To characterize the bifurcation, we measure the temperature
at the end of the tubular reactor ($\theta$), and introduce
the following indicator of stability.
\begin{equation*}
 \label{eq:indicator}
 g(D) =
 \left\{
  \begin{array}{ll}
  \alpha(D), & \text{for decaying oscillations,} \\
  (\gamma r(D))^2, & \text{for limit cycle oscillations.}
 \end{array} \right.
\end{equation*}
$\alpha$ is the growth rate, estimated by fitting the
temperature in the last two cycles of oscillation to the
approximation
$\theta \approx \theta_0 + \bar{\theta} e^{\alpha t}$, where
$t$ denotes time.
Furthermore, $r$ is the amplitude of limit cycle
oscillations, and $\gamma = 25$ is a parameter that controls
the intensity of the chemical reaction.

Our goal is to locate the critical Damk\"ohler number using
two numerical models of the tubular reactor dynamics.
The first model results from a centered finite-difference
discretization of the governing equations and boundary
conditions, and corresponds to IS0.
The second model is a reduced-order model based on the 
combination of proper orthogonal decomposition and the
discrete empirical interpolation method, and corresponds to
IS1.
Both models are described in details by
Zhou~\cite{zhou:2010}.

\begin{figure}[b!]
 \begin{center}
  \includegraphics[scale=0.45,
                   trim={0 0.65in 0 0.05in},
                   clip]{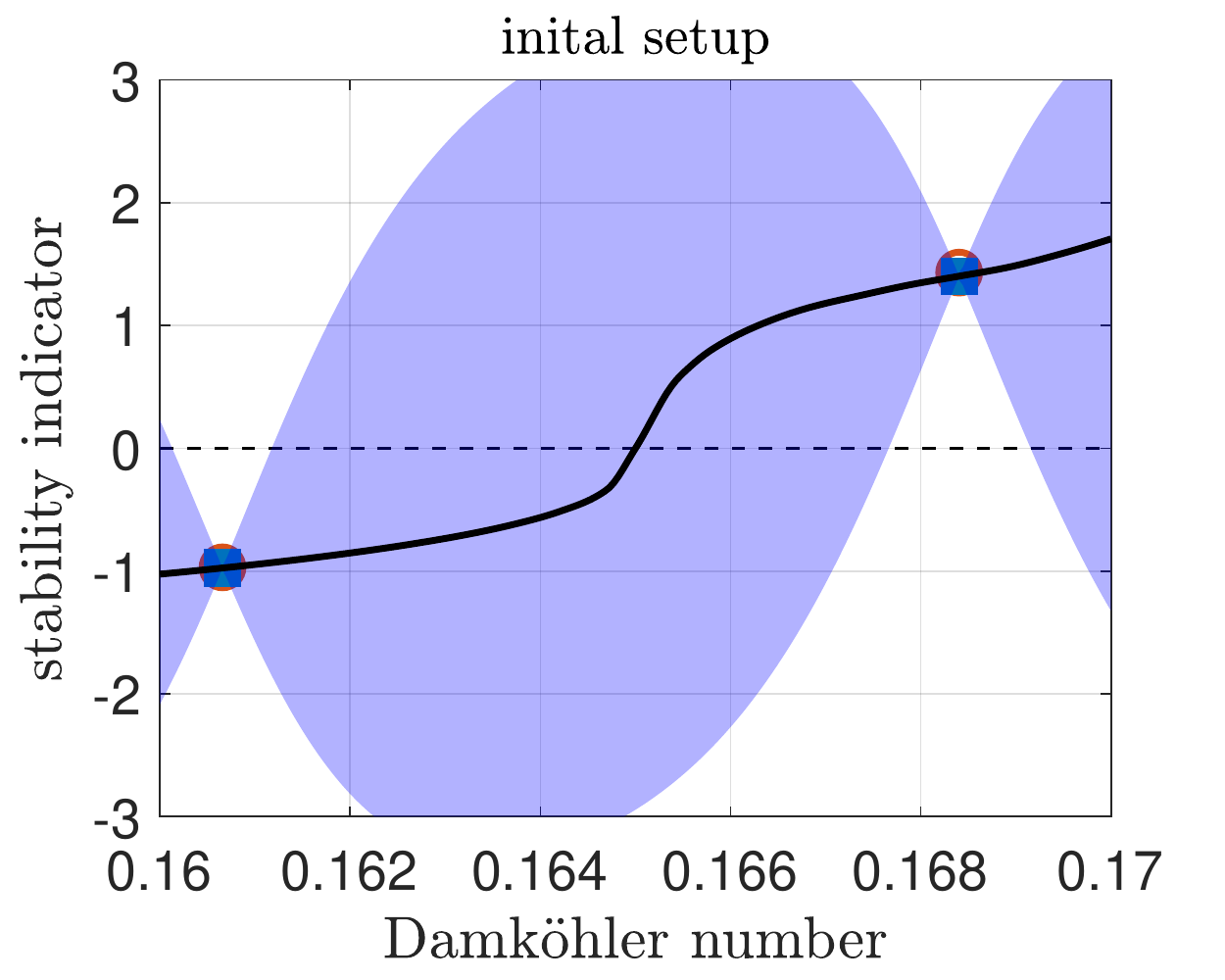}
  \includegraphics[scale=0.45,
                   trim={0.6in 0.65in 0 0.05in},
                   clip]{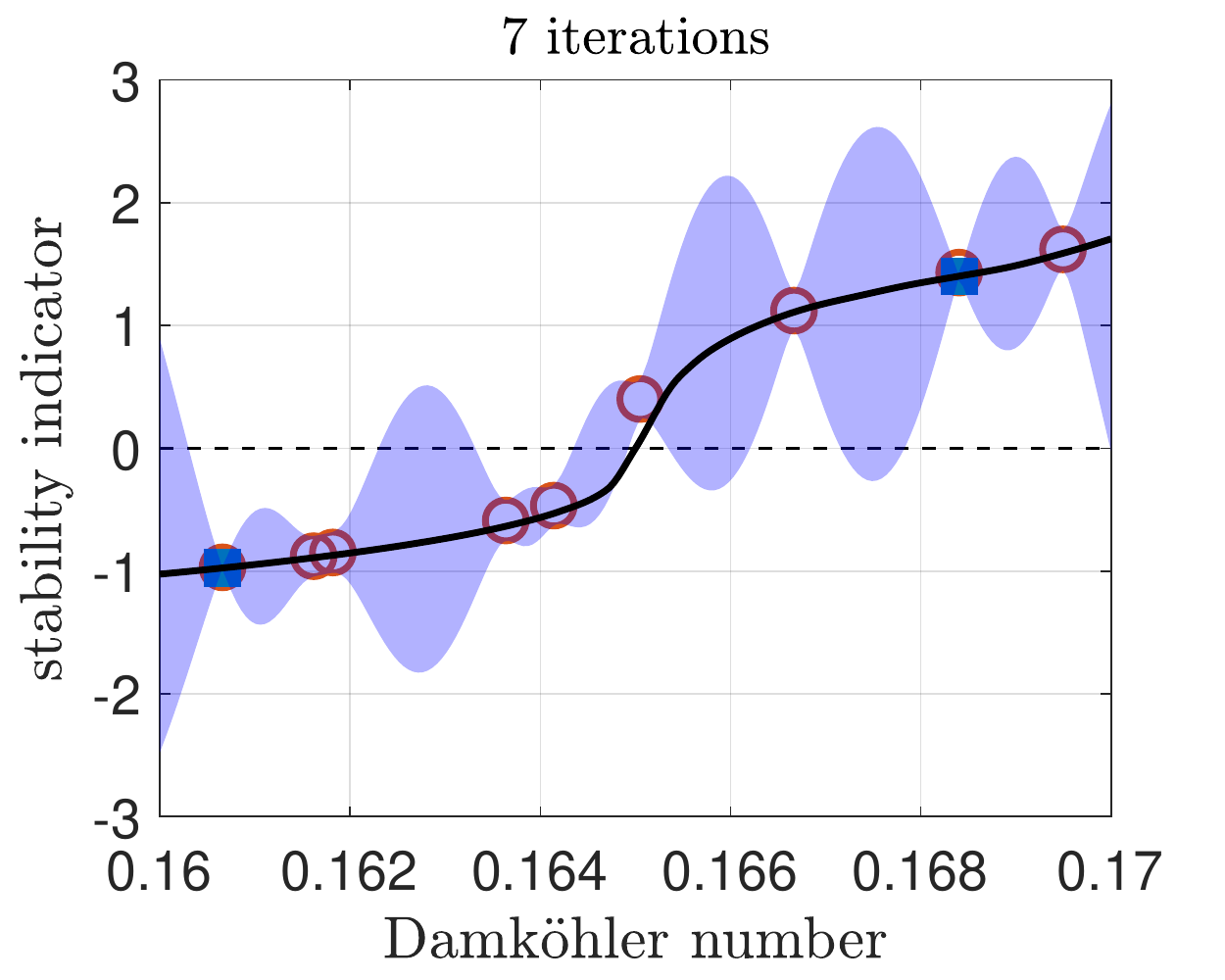} \\[5pt]
  \includegraphics[scale=0.45,
                   trim={0 0.05in 0 0.05in},
                   clip]{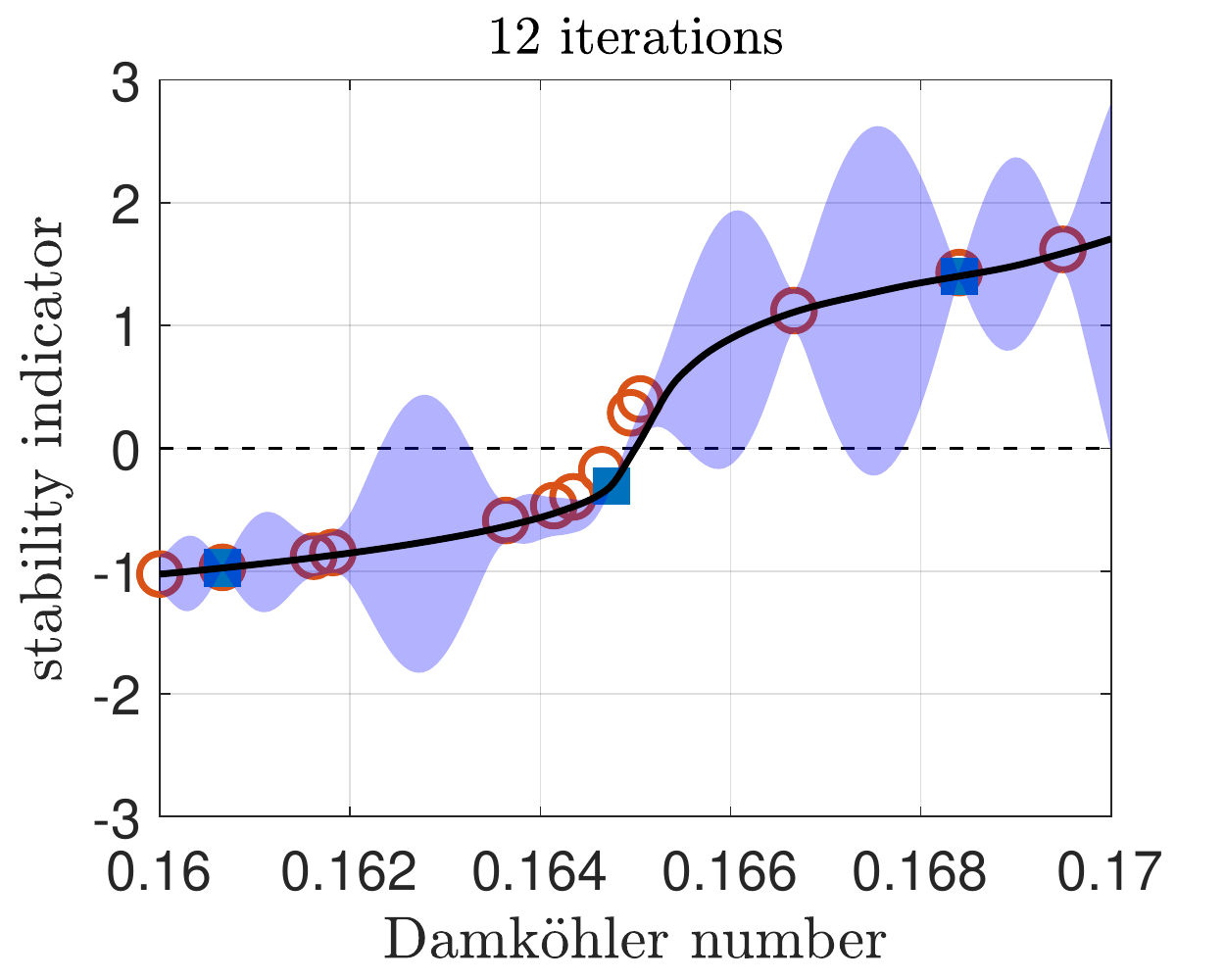}
  \includegraphics[scale=0.45,
                   trim={0.6in 0.05in 0 0.05in},
                   clip]{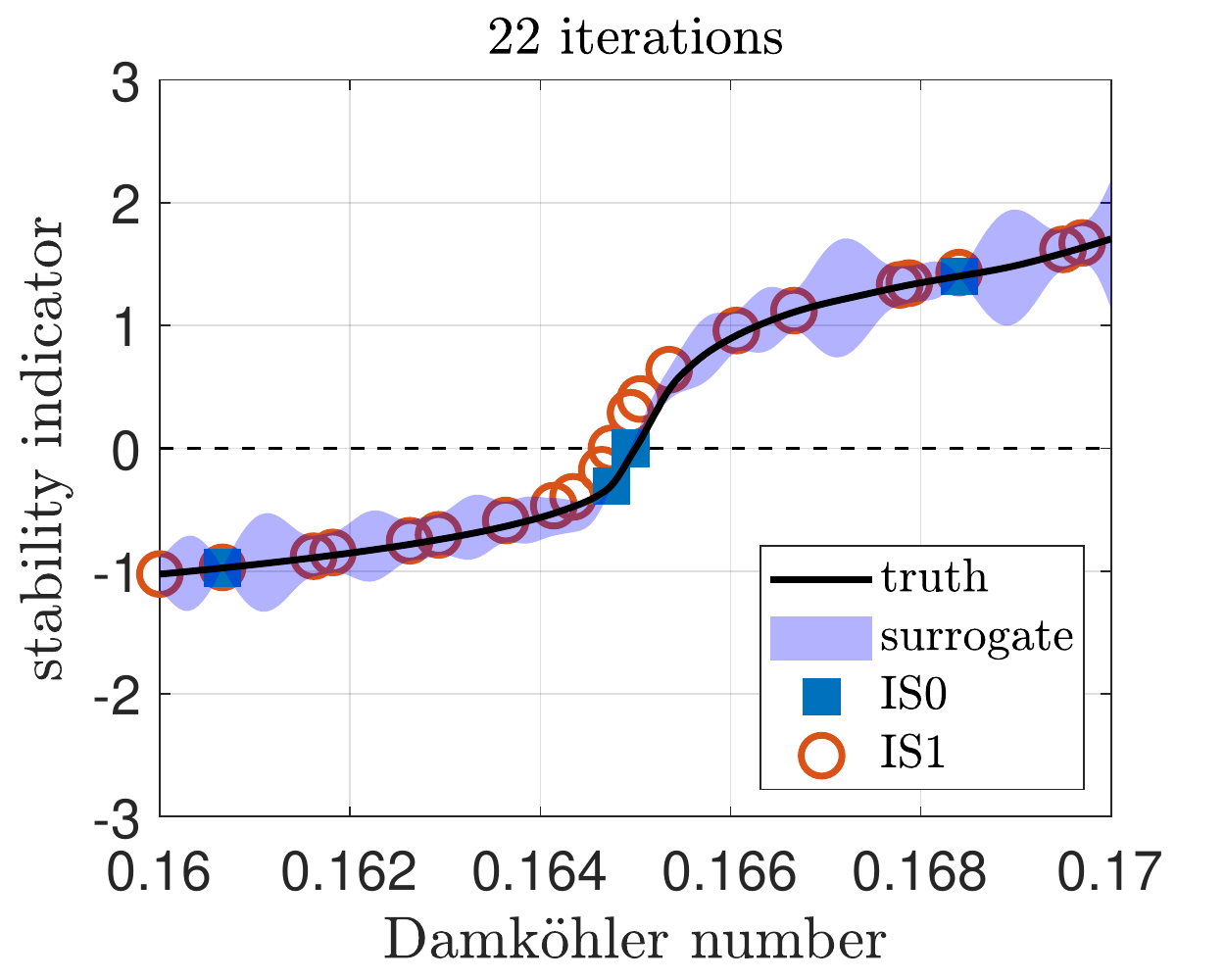}
 \end{center}
 \caption{Locating the H\"opf bifurcation of a tubular
          reactor (zero contour of stability indicator).
          Shaded area: $\pm 3\sigma$ around the
          mean of the GP surrogate.
          CLoVER locates the bifurcation after 22
          iterations, using only 4 evaluations of IS0.}
 \label{fig:tubular_reactor_snapshots}
\end{figure}

\begin{figure}[t!]
 \begin{center}
  \includegraphics[scale=0.4,
                   trim={0, 0.05in, 0, 0.05in},
                   clip]{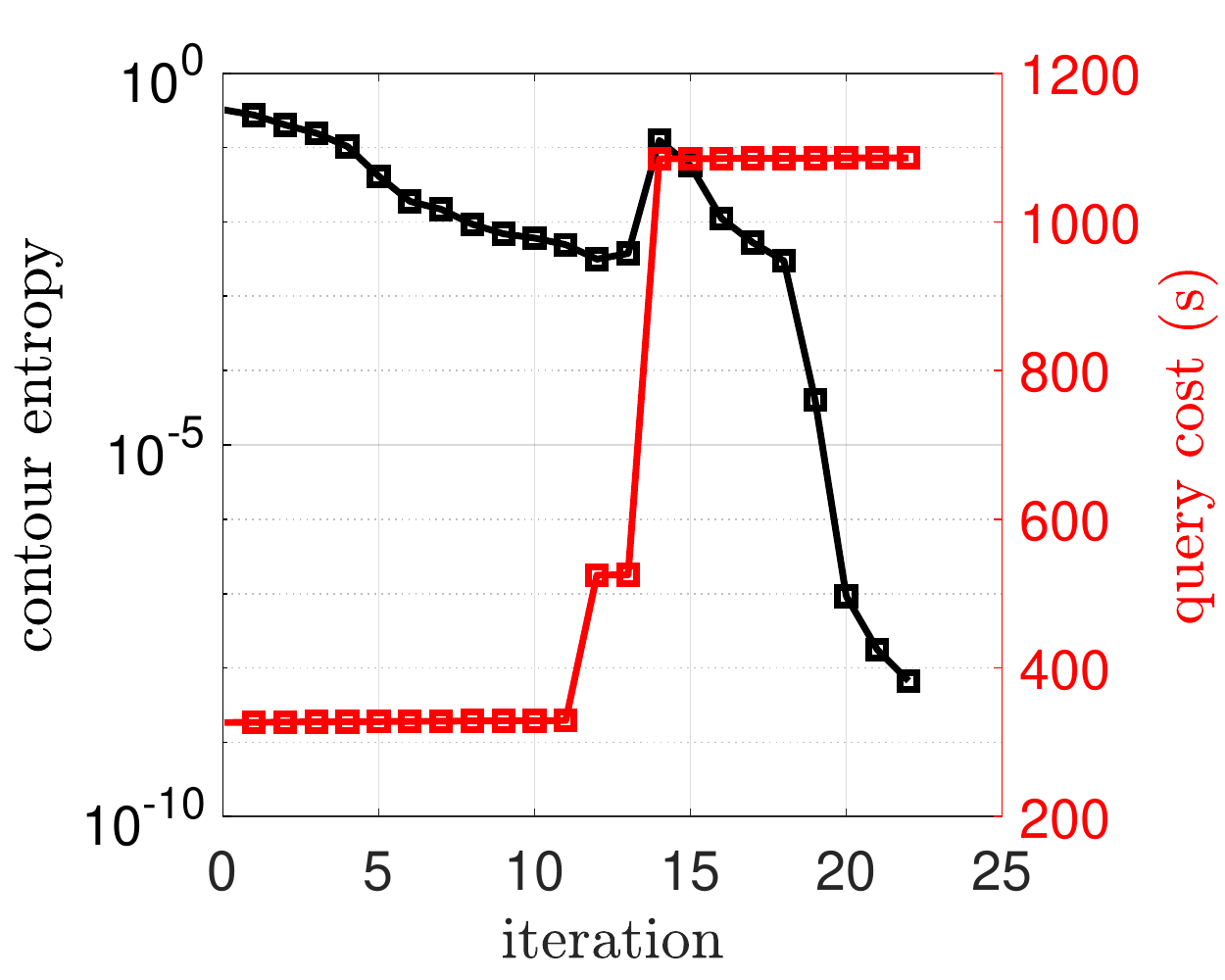}
  \hspace{0.2in}
  \includegraphics[scale=0.4,
                   trim={0, 0.05in, 0, 0.05in},
                   clip]{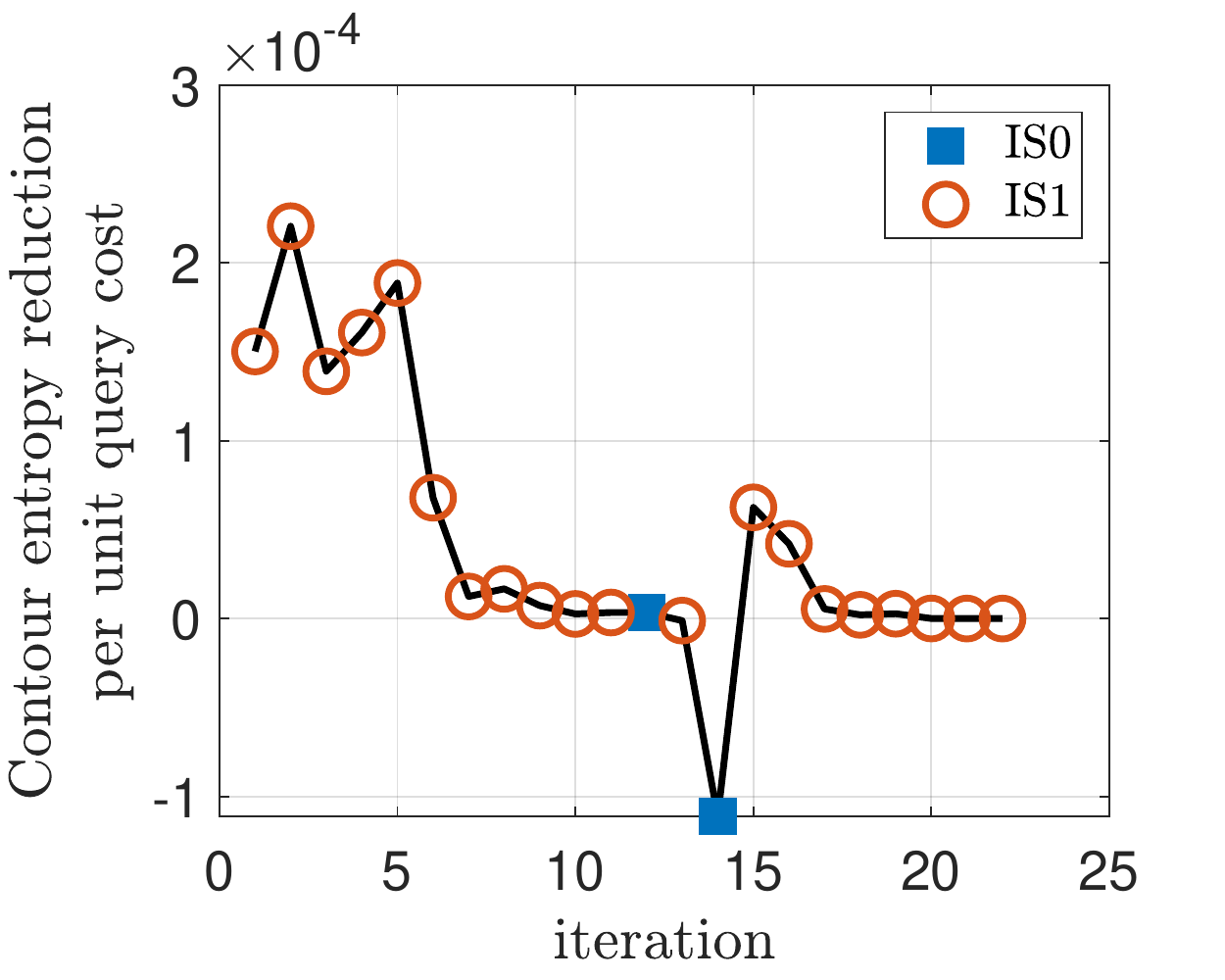}
 \end{center}
 \caption{Left: Contour entropy and query cost during the
          iterations of the CLoVER algorithm.
          Right: Reduction in contour entropy per unit query
          cost at every iteration.
          CLoVER explores IS1 to decrease the uncertainty
          about the location of the bifurcation before using
          evaluations of expensive IS0.}
 \label{fig:tubular_reactor_H}
\end{figure}

Figure~\ref{fig:tubular_reactor_snapshots} shows the samples
selected by CLoVER, and the uncertainty predicted by the
GP surrogate at several iterations.
The algorithm starts with two random evaluations of both
models.
This information is used to compute a MAP estimate of the
hyperparameters of the GP surrogate, using the procedure
recommended by Poloczek et al.~\cite{poloczek:2017}%
\footnote{%
For the length scales of the covariance kernels, Poloczek et
al.~\cite{poloczek:2017} recommend using normal distribution
priors with mean values given by the range of \domain\/ in
each coordinate direction.
We found this heuristics to be only appropriate for
functions that are very smooth over \domain\/.
In the present example we adopt $d_0 = 0.002$ and
$d_1 = 0.0005$ as the mean values for the length scales of
$\Sigma_0$ and $\Sigma_1$, respectively.}
and to provide an initial estimate of the surrogate.
In this example we use covariance kernels of the Mat\'ern
class~\cite{rasmussen:2005} with $\nu = 5/2$, and zero mean
functions.
After these two initial evaluations, CLoVER explores the
parameter space using 11 evaluations of IS1.
This behavior is expected, since the query cost of IS0 is
500-3000 times the query cost of IS1.
Figure~\ref{fig:tubular_reactor_H} shows the evolution of
the contour entropy and query cost along the iterations.
After an exploration phase, CLoVER starts exploiting near
$D = 0.165$.
Two evaluations of IS0, at iterations 12 and 14, allow
CLoVER to gain confidence in predicting the critical
Damk\"ohler number at $D_{cr} = 0.165$.
After eight additional evaluations of IS1, CLoVER determines
that other bifurcations are not likely in the parameter
range under consideration.
CLoVER concludes after a total of 22 iterations, achieving
$\lse = 6 \times 10^{-9}$.

\subsection{Comparison between CLoVER and existing
            algorithms for single information source}
\label{sub:comparison}

Here we compare the performance of CLoVER with a single
information source to those of algorithms
EGRA~\cite{bichon:2008}, Ranjan~\cite{ranjan:2008},
TMSE~\cite{picheny:2010}, TIMSE~\cite{bect:2012},
and SUR~\cite{bect:2012}.
This comparison is based on locating the contour $g = 80$ of
the two-dimensional Branin-Hoo function~\cite{bingham:2018}
within the domain $\domain = [-5, 10] \times [0, 15]$.
We discuss a similar comparison, based on a different
problem, in the supplementary material.

The algorithms considered here are implemented in the R
package \texttt{KrigInv}~\cite{chevalier:2014}.
Our goal is to elucidate the effects of the distinct
acquisition functions, and hence we execute \texttt{KrigInv}
using the same GP prior and schemes for optimization and
integration as the ones used in CLoVER.
Namely, the GP prior is based on a constant mean function
and a squared exponential covariance kernel, and the
hyperparameters are computed using MLE.
The integration over \domain\/ is performed with the
trapezoidal rule on a $50 \times 50$ uniform grid, and
the optimization set $\mathcal{A}$ is composed of a
$30 \times 30$ uniform grid.
All algorithms start with the same set of 12 random
evaluations of $g$, and we repeat the computations 100
times using different random sets of evaluations for
initialization.

\begin{figure}[h!]
 \begin{center}
  \includegraphics[scale=0.45,
                   trim={0, 0.05in, 0, 0},
                   clip]{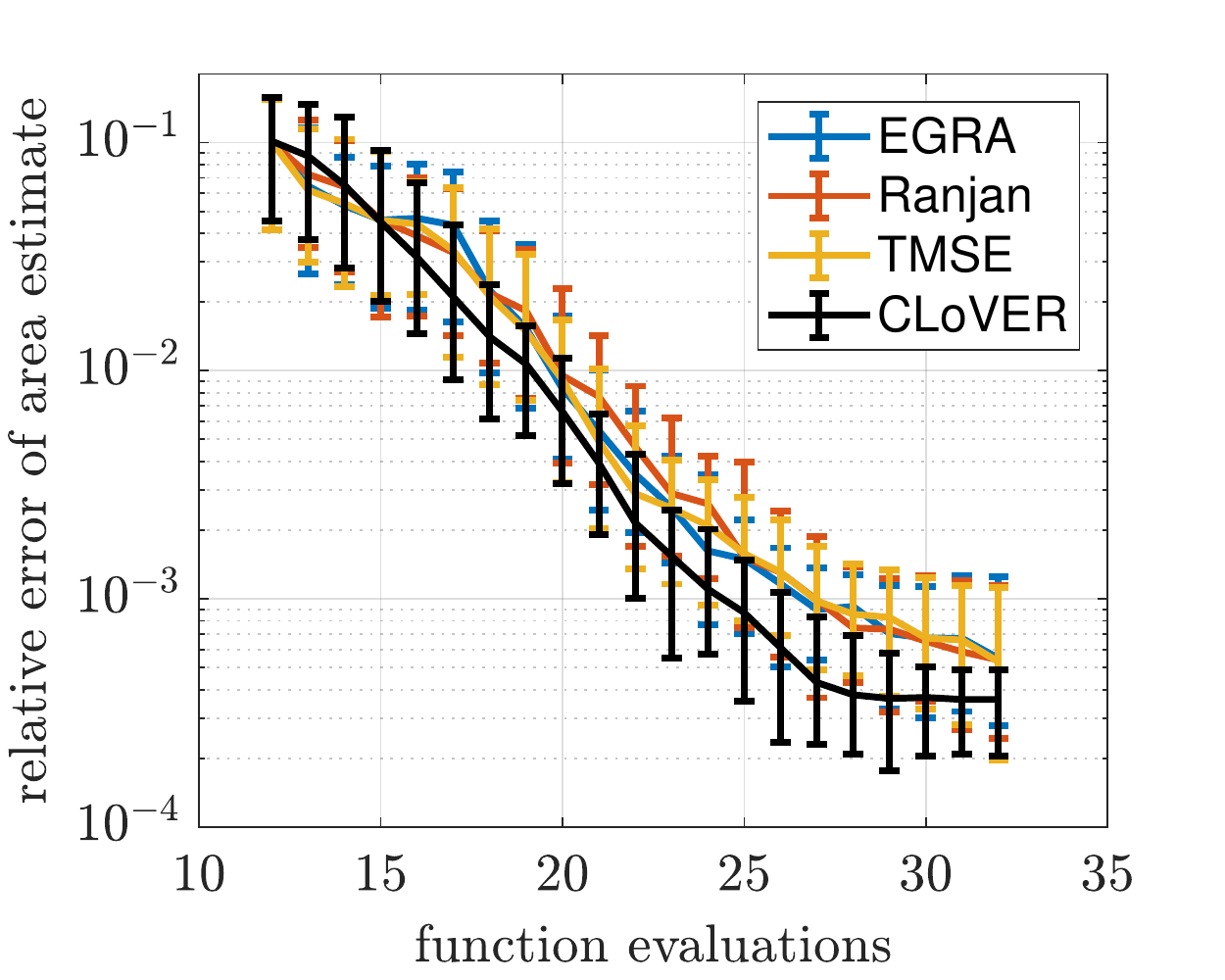}
  \hspace{0.2in}
  \includegraphics[scale=0.45,
                   trim={0, 0.05in, 0, 0},
                   clip]{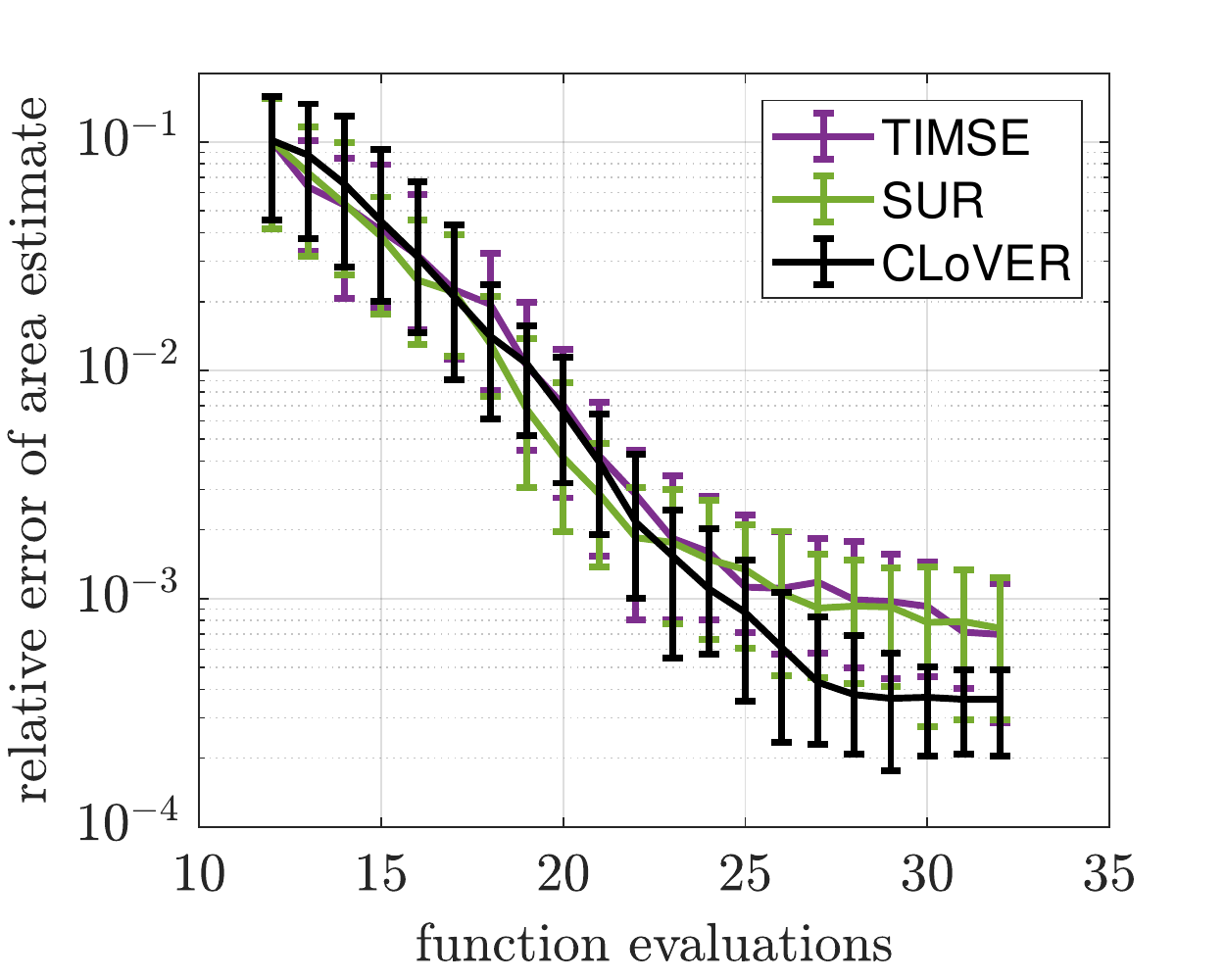}
 \end{center}
 \caption{Relative error in the estimate of the area of the
          set $S$ (median, 25th, and 75th percentiles).
          Left: comparison between CLoVER and greedy
                algorithms EGRA, Ranjan, and TMSE.
          Right: comparison between CLoVER and one-step
                 look ahead algorithms TIMSE and SUR.}
 \label{fig:comparison}
\end{figure}

We compare performance by computing the area of the set
$S = \{ \x \in \domain \mid g(\x) > 80 \}$.
We compute the area using Monte Carlo integration with
$10^6$ samples, and compare the results to a reference
value computed by averaging 20 Monte Carlo estimates based
on evaluations of $g$: $\text{area}(S) = 57.8137$.
Figure~\ref{fig:comparison} compares the relative error in
the area estimate computed with the different algorithms.
All algorithms perform similarly, with CLoVER achieving a
smaller error on average.

\subsubsection*{Acknowledgments}

This work was supported in part by the U.S.\ Air Force Center of Excellence on Multi-Fidelity Modeling of Rocket Combustor Dynamics, Award FA9550-17-1-0195, and by the AFOSR MURI on managing multiple information sources of multi-physics systems, Awards FA9550-15-1-0038 and FA9550-18-1-0023.

\bibliographystyle{ieeetr}
\bibliography{references}

\clearpage

\setcounter{page}{1}

\begin{center}
{\huge \textbf{Contour location via entropy reduction \\ leveraging multiple information sources}}\\[10pt]
{\large \textbf{Alexandre N.\ Marques \hspace{0.2in}
Remi R.\ Lam \hspace{0.2in}
Karen E.\ Willcox}\\[15pt]
\textit{Supplementary material}}
\end{center}

\renewcommand*{\thesection}{\Alph{section}}
\setcounter{section}{0}

\renewcommand{\thefigure}{S.\arabic{figure}}


\section{Additional comparison between CLoVER and existing
         algorithms for single information source}
\label{sec:comparison}

Here we compare the performance of CLoVER with a single
information source to those of the algorithms EGRA~[15],
Ranjan~[16], TMSE~[17], TIMSE~[18], and SUR~[18], similarly
to case discussed in Sect.~4.3.
In this investigation, we solve the problem described in
example 1 of Ref.~[15] (multimodal function).
Consider the random variable $\x \sim
\mathcal{N}(\boldsymbol{\mu}_x, \boldsymbol{\Sigma}_x)$,
where
\begin{align*}
 \boldsymbol{\mu}_x &=
 \begin{Bmatrix} 1.5\\ 2.5 \end{Bmatrix}, &
 \boldsymbol{\Sigma}_x &= 
 \begin{bmatrix} 1 & 0 \\ 0 & 1 \end{bmatrix},
\end{align*}
in the domain $\domain = [-4, 7] \times [-3, 8]$.
The goal is to estimate the probability
$p_f = P(g(\x) > 0)$, where
\begin{equation*}
 g(\x) = \dfrac{(x_1^2 + 4)(x_2 - 1)}{20}
       - \sin \left( \dfrac{5 x_1}{2} \right) - 2.
\end{equation*}
We estimate the probability $p_f$ by first locating the zero 
contour of $g$ and then computing a Monte Carlo integration
based on the surrogate model:
\begin{equation*}
 p_f \approx \dfrac{1}{N}\sum_{i=1}^N \mathbb{I}_f (\x_i),
 \quad \x \sim
 \mathcal{N}(\boldsymbol{\mu}_x, \boldsymbol{\Sigma}_x).
\end{equation*}
where
\begin{equation*}
 \mathbb{I}_f(\x_i) =
 \left\{ \begin{array}{ll}
  1, & \mu(0, \x_i) > 0, \\
  0, & \text{otherwise},
 \end{array} \right.
\end{equation*}
$\mu(0, \x_i)$ denotes the mean of $f(0, \x_i)$, and
$N = 10^6$ is the number of Monte Carlo samples.
We assess the accuracy of the estimates by comparing them to
a reference value computed by averaging 20 Monte Carlo
estimates based on evaluations of $g$: $p_f = 0.03133$.

The R package \texttt{KigInv}~[19] provides implementations
of the algorithms listed above.
As in Sect.~4.3, we execute \texttt{KrigInv} using the same
GP prior and schemes for optimization and integration as the
ones used in CLoVER.
Namely, the GP prior is based on a constant mean function
and a squared exponential covariance kernel, and the
hyperparameters are computed using MLE.
The integration over \domain\/ is performed with the
trapezoidal rule on a $50 \times 50$ uniform grid, and
the optimization set $\mathcal{A}$ is composed of a
$30 \times 30$ uniform grid.
All algorithms start with the same set of 10 random
evaluations of $g$, and stop when the acquisition function
reaches a value of $10^{-8}$ or after 50 function
evaluations, whichever occurs first.
We repeat the computations 100 times using different
random sets of evaluations for initialization. 

\begin{figure}[t!]
 \begin{center}
  \includegraphics[scale=0.5]
                  {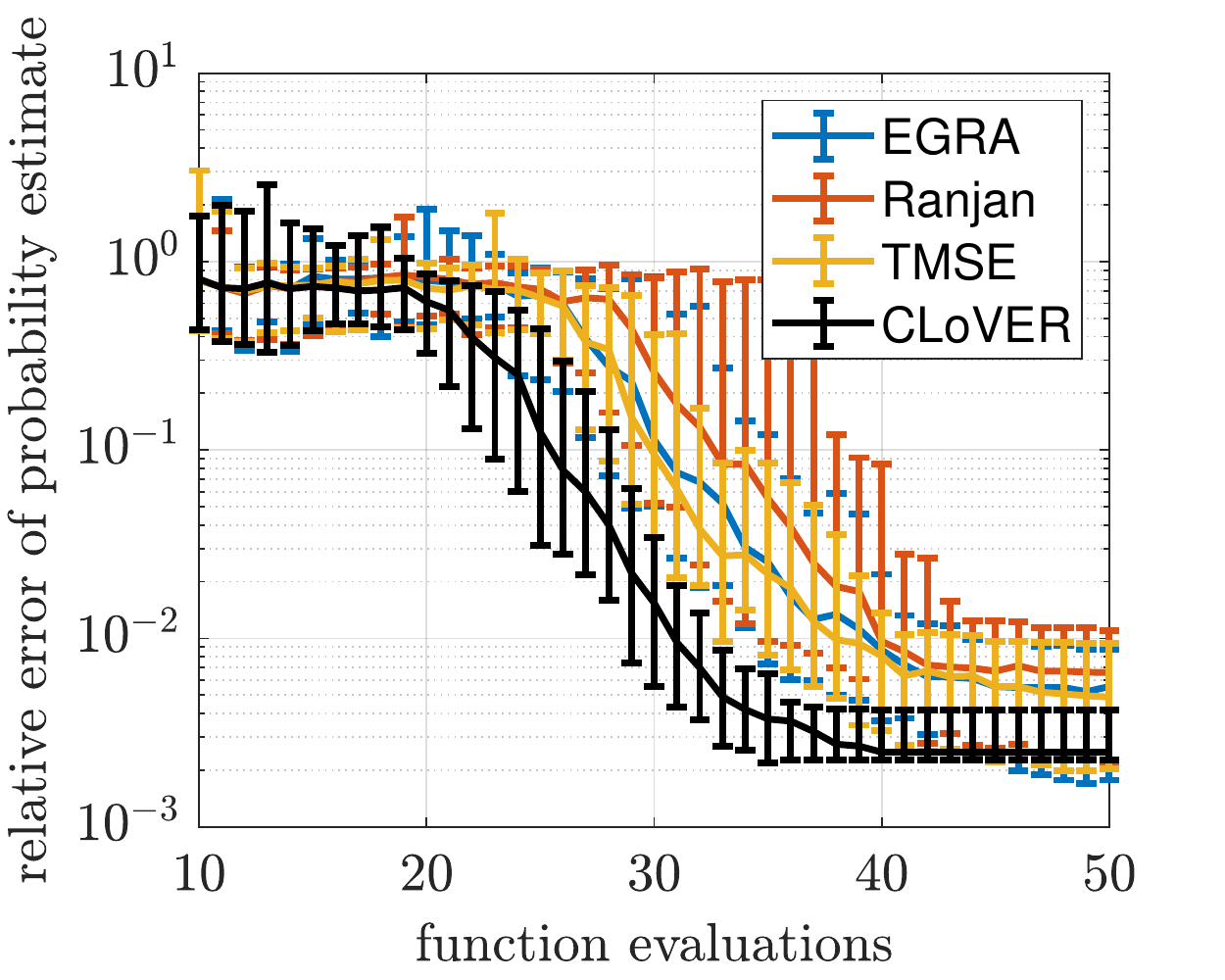}
  \includegraphics[scale=0.5]
                  {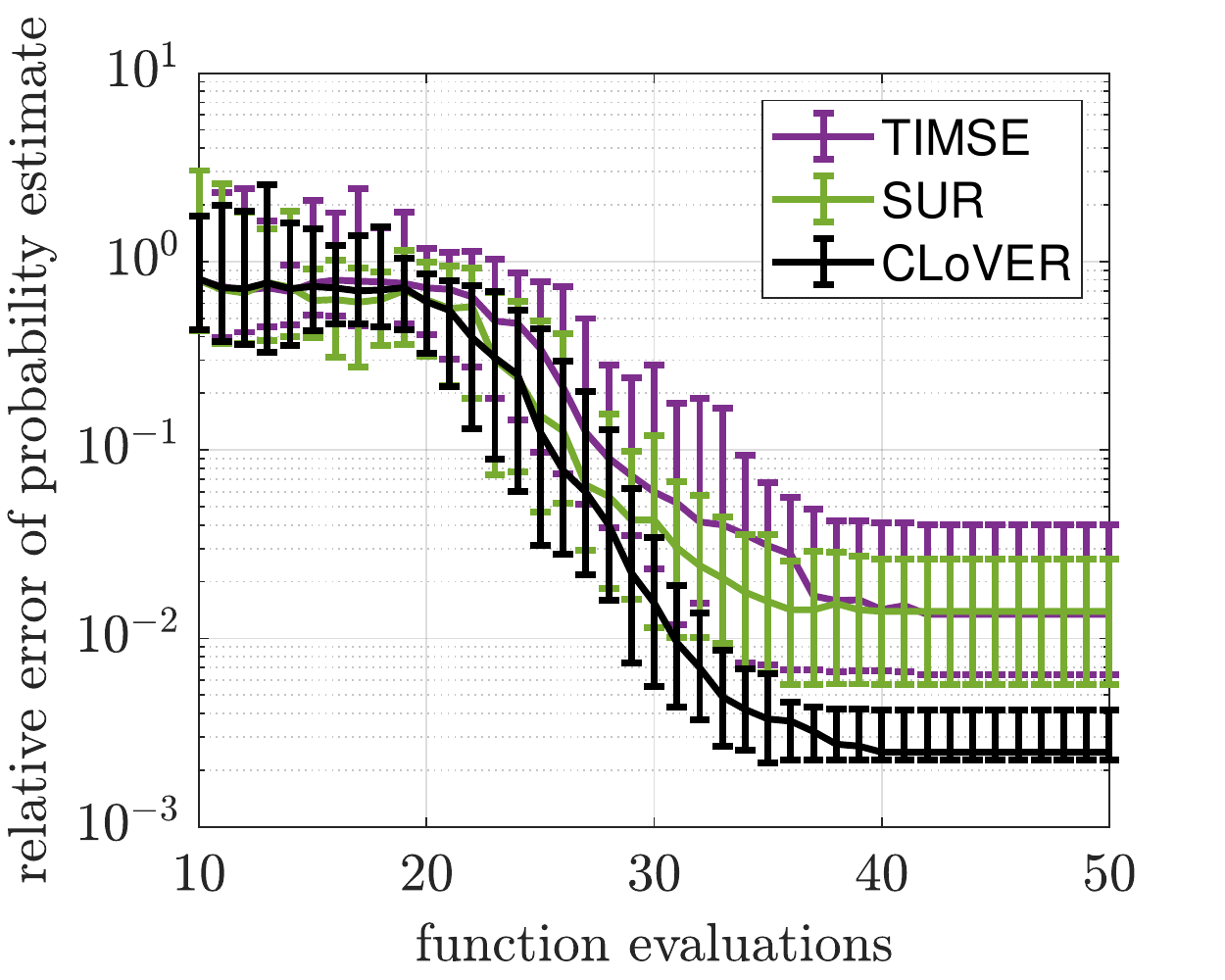}
 \end{center}
 \caption{Relative error in the estimate of the
          probability $p_f$ (median, 25th, and 75th
          percentiles).
          Left: comparison between CLoVER and greedy
                algorithms EGRA, Ranjan, and TMSE.
          Right: comparison between CLoVER and one-step
                 look ahead algorithms TIMSE and SUR.}
 \label{fig:sup:pf}
\end{figure}

Figure~\ref{fig:sup:pf} shows the relative error of the
estimates of $p_f$.
We observe that on average CLoVER results in a faster error
decay, and converges to lower error level.
The median number of function evaluations are the following.
CLoVER: 38, EGRA: 42, Ranjan: 42, TMSE: 41, TIMSE: 50,
SUR:~33.

\begin{figure}[t!]
 \begin{center}
  \includegraphics[scale=0.5]
                  {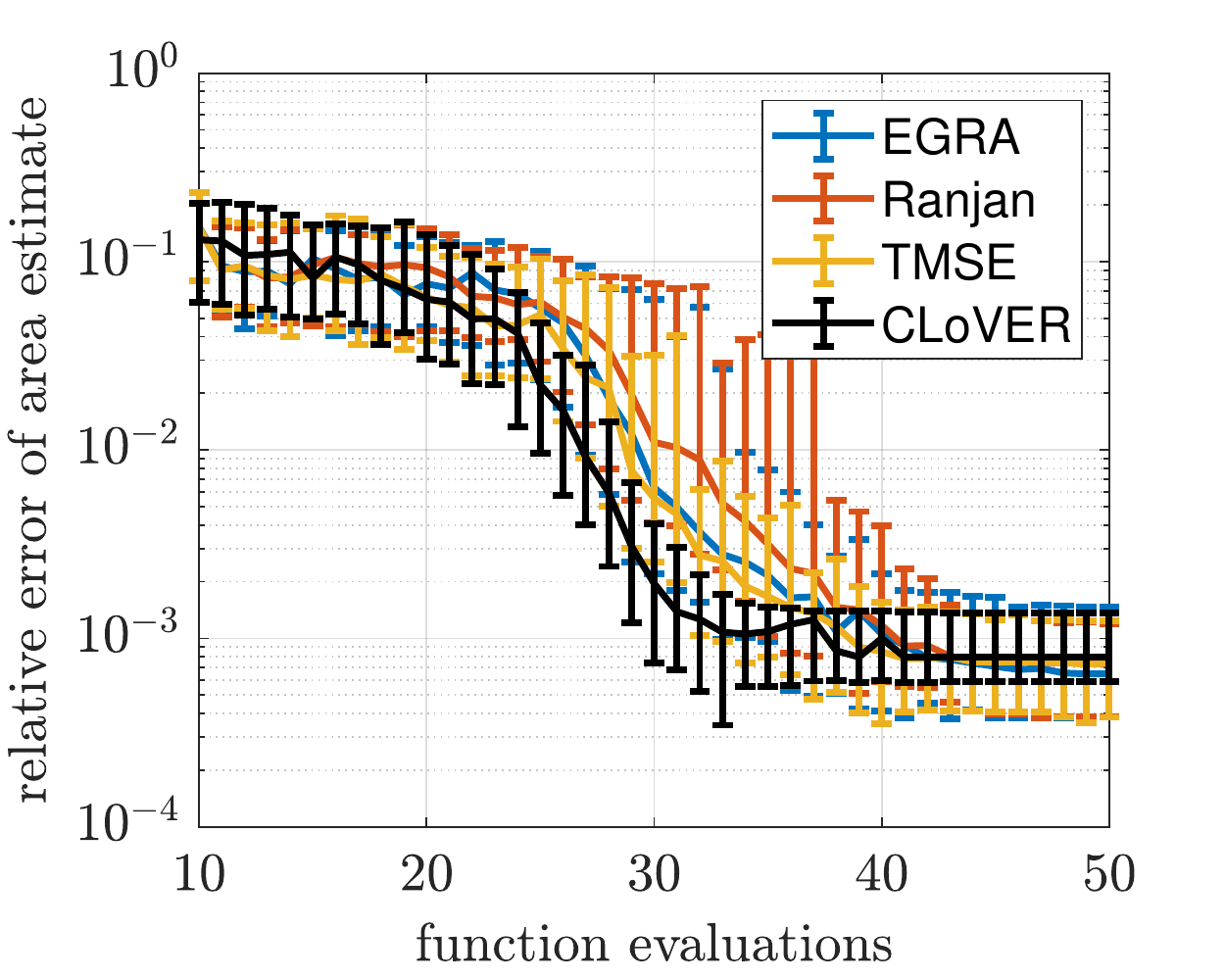}
  \includegraphics[scale=0.5]
                  {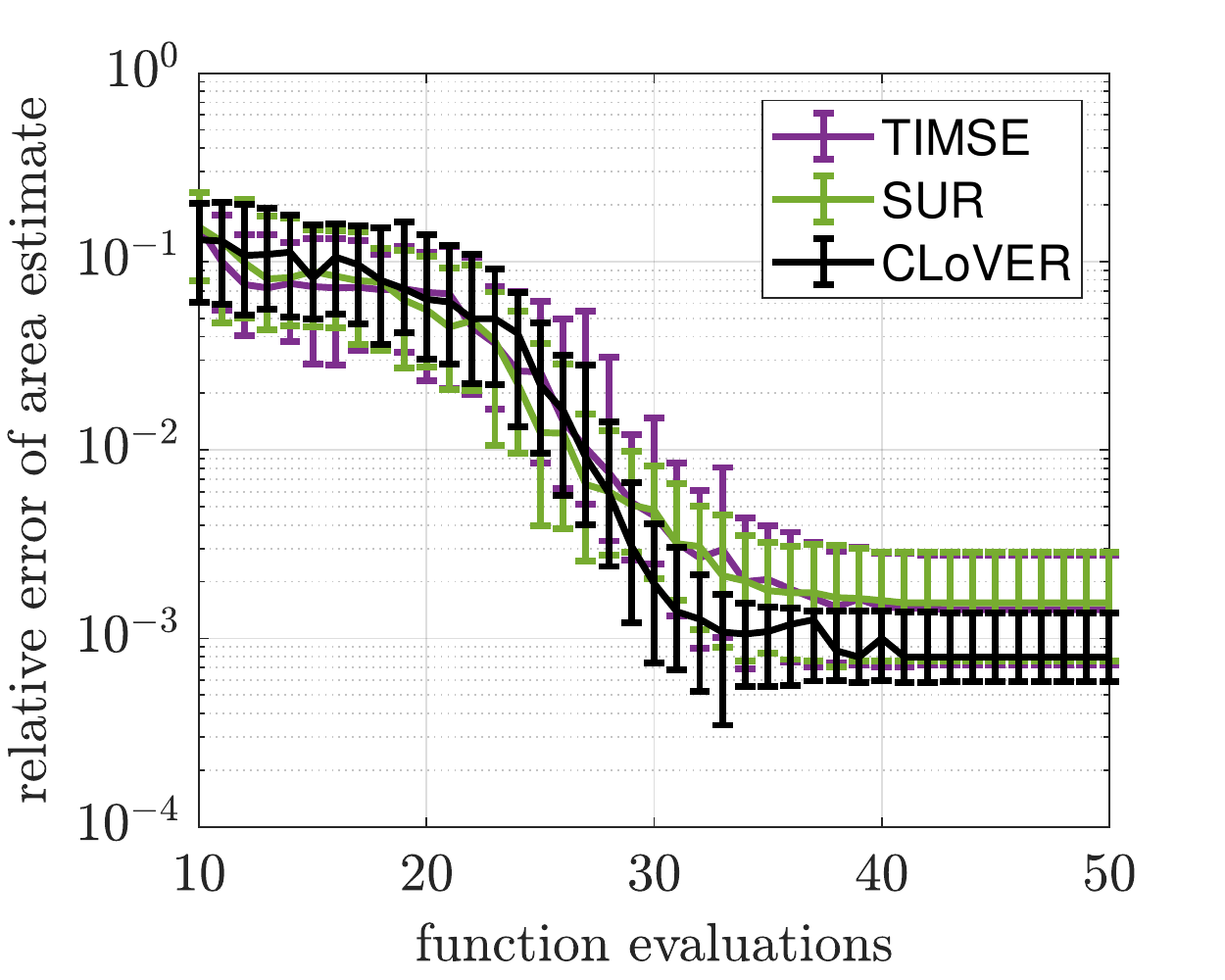}
 \end{center}
 \caption{Relative error in the estimate of the area of set
          $S$ (median, 25th, and 75th percentiles).
          Left: comparison between CLoVER and greedy
                algorithms EGRA, Ranjan, and TMSE.
          Right: comparison between CLoVER and one-step
                 look ahead algorithms TIMSE and SUR.}
 \label{fig:sup:area}
\end{figure}

We also evaluate the algorithms by computing the area of
the subdomain $S = \{ \x \in \domain \mid g(\x) > 0 \}$
(shaded area in Figure~3).
We estimate the area using Monte Carlo integration with
$10^6$ samples in the region $[-4, 7] \times [1.4, 8]$, and
compare the results to a reference value computed by
averaging 20 Monte Carlo estimates based on evaluations of
$g$: $\text{area}(S) = 36.5541$.
Figure~\ref{fig:sup:area} shows the relative error in the
estimates of the area of the set $S$.
CLoVER also presents a faster decay of the error of the
area estimate.

\section{Trade-off between exploration and exploitation}
\label{sec:epsilon}

As disussed in Sect.~3, the concept of contour entropy uses
the parameter $\epsilon\/$ as a tolerance in the definition
of the zero contour.
This parameter also provides a control over the trade-off
between exploration (sampling in regions where uncertainty
is large) and exploitation (sampling in regions of
relatively low uncertainty, but likely close to the zero
contour).
An algorithm that favors exploration may be ineffecient
because it evaluates many samples in regions far from the
zero contour, whereas an algorithm that favors exploitation
may fail to identify disjoint parts of the contour because 
it concentrates samples on a small region of the domain.
In general, larger values of $\epsilon\/$ result in more
exploration than exploitation, and vice-versa.

\begin{figure}[h!]
  \begin{center}
    \includegraphics[width=0.5\textwidth]
                    {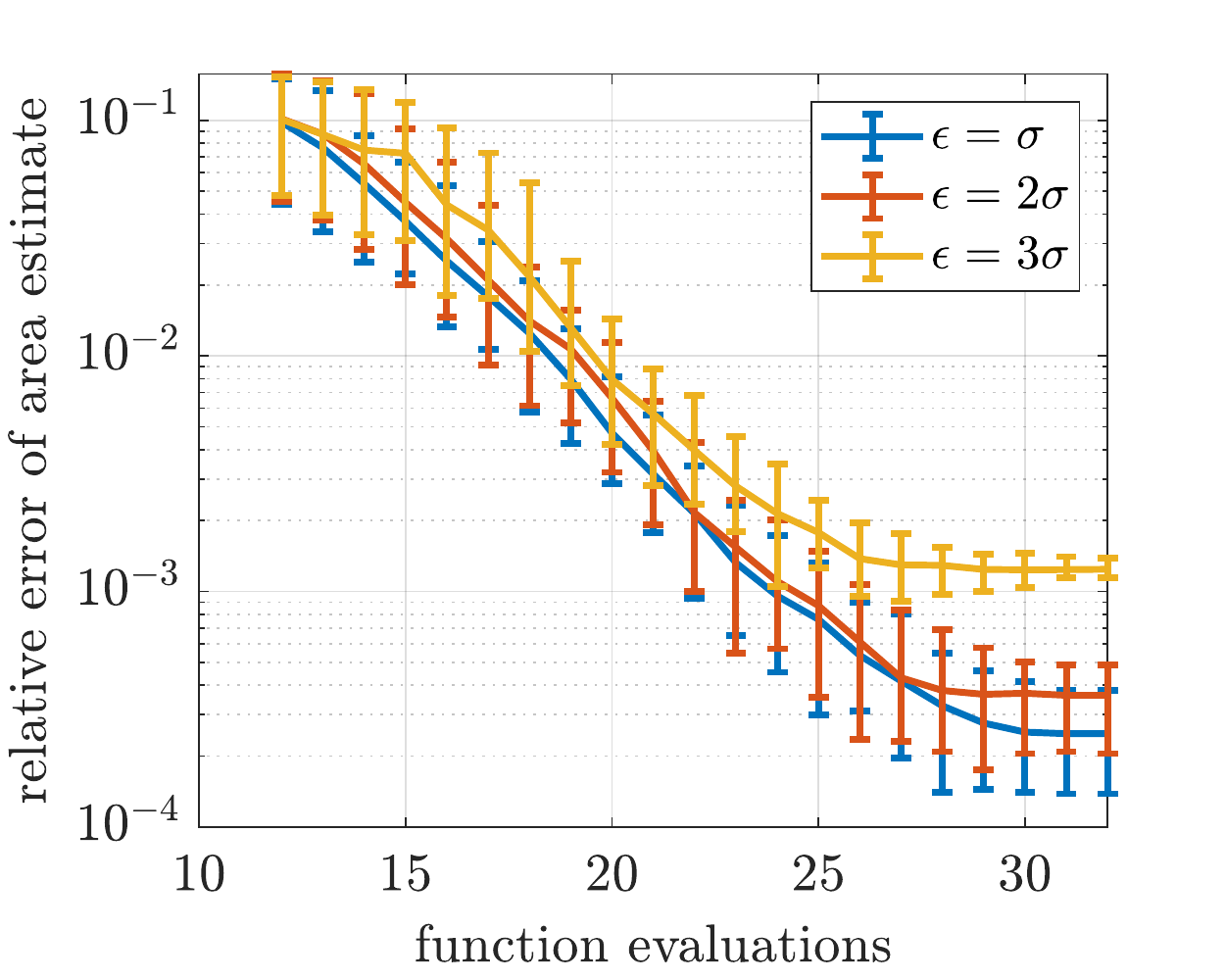}
  \end{center}
  \caption{Influence of $\epsilon$ in the convergence of
           CLoVER.
           The plot shows the relative error (median, 25th
           and 75th percentiles) in the area of the set
           $S$.}
 \label{fig:epsilon}
\end{figure}

We find that making $\epsilon\/$ proportional to the
standard deviation of the surrogate model,
\begin{equation*}
 \epsilon(\x) = \ce \sigma(\x),
\end{equation*}
provides a good balance between exploration and
exploitation.
To determine the constant of proportionality \ce\, we
repeat the experiment described in Sect.~4.3 with
$\ce \in \{1, 2, 3\}$.
We measure the accuracy in the prediction of the zero
contour by computing the area of the excursion set
$S = \{ \x \in [-5, 10] \times [0, 15] \mid
g(\x) > 80 \}$, where $g$ denotes the two-dimensional
Branin-Hoo function~[26].
Figure~\ref{fig:epsilon} shows the convergence in the
relative error in the estimates of $S$ computed with
different values of \ce\/.

We observe that in all cases CLoVER identified the zero
contour, although with varying levels of accuracy.
As expected, $\ce = 1$ leads to a more exploitative
algorithm that on average converges faster to the zero
contour.
Choosing $\ce = 2$ does not affect the convergence rate
significantly, but leads to a slightly larger error in the
area estimate.
Finally, setting $\ce = 3$ considerably degrades the
performance of the algorithm.

Although we do not observe adverse effects of choosing
$\ce = 1$ in this particular example, we also do not observe
significant improvements with respect to $\ce = 2$.
For this reason, we favor choosing $\ce = 2$ to avoid an
excessively exploitative algorithm.
We find this heuristic to work well in general problems.


\section{Computational cost}
\label{sec:cost}

The computational cost of CLoVER is comparable to that of
other algorithms with one-step look ahead acquisition
functions (e.g., TIMSE and SUR~[18]).
Greedy acquisition functions are cheaper to evaluate, but
offer no natural form of selecting information sources when
more than one is available.
In addition, Chevalier et al.~[19] report that one-step
look ahead strategies are more efficient in selecting
samples because they take into account global effects of new
observations, resulting in a lower number of function
evaluations for comparable accuracy.
Most importantly, the multi-information source setting
considered in this paper is relevant when the highest
fidelity information source is expensive.
In this scenario, the cost of selecting new samples
($\sim 4s$ for a two-dimensional problem) is normally small
in comparison to function evaluations.

The computational cost of CLoVER is dominated by evaluating
the variance $\bar{\sigma}^2$ within the integral of
Eq.~(10) (see Sect.~3.3).
In general, the cost of evaluating the variance of a GP
surrogate after $n$ observations is \ord{n^3}.
(The cost can be reduced to \ord{n \log^2 n} for specific
covariance functions, and large values of $n$).
Therefore, the total computations cost scales as
\ord{n_a n_i n^3}, where $n_a$ denotes the number of points
in the optimization set $\mathcal{A}$, and $n_i$ denotes the
number of points used to evaluate the integral of Eq.~(10).

\end{document}